\documentclass[]{bytedance_seed}

\usepackage[toc,page,header]{appendix}

\usepackage{amsmath,amsfonts,bm}

\def\eqref#1{equation~\ref{#1}}

\def\1{\bm{1}}

\DeclareMathAlphabet{\mathsfit}{\encodingdefault}{\sfdefault}{m}{sl}
\SetMathAlphabet{\mathsfit}{bold}{\encodingdefault}{\sfdefault}{bx}{n}

\usepackage{url}            
\usepackage{amsfonts}       
\usepackage{nicefrac}       
\definecolor{oursblue}{HTML}{D6E6F5}
\hypersetup{breaklinks=true}
\DeclareTextFontCommand{\textbf}{\bfseries} 

\newcommand{\genimg}[2]{\includegraphics[height=#1]{figs/gen_main/#2.jpg}}

\definecolor{qualA}{HTML}{E41A1C}
\definecolor{qualB}{HTML}{377EB8}
\definecolor{qualC}{HTML}{4DAF4A}
\newcommand{\qtok}[1]{\texttt{\textless#1\textgreater}}
\newcommand{\qA}[1]{\textcolor{qualA}{\textbf{#1}}}
\newcommand{\qB}[1]{\textcolor{qualB}{\textbf{#1}}}
\newcommand{\qC}[1]{\textcolor{qualC}{\textbf{#1}}}
\newcommand{\qualbox}[1]{\fbox{\parbox{\dimexpr\linewidth-2\fboxsep-2\fboxrule\relax}{\scriptsize\setlength{\parskip}{2pt}#1}}}

\usepackage{amssymb}
\usepackage{graphicx}
\usepackage{wrapfig}
\usepackage{placeins}
\usepackage{needspace}
\usepackage{etoc}
\usepackage{enumitem}
\usepackage{subcaption}
\usepackage{colortbl}
\usepackage{amsmath,amssymb}
\usepackage{fontawesome5}
\newcommand{\emailicon}{\text{\faEnvelope[regular]}}
\usepackage[ruled,linesnumbered]{algorithm2e}


\usepackage{multirow} 
\newcommand{\tablestyle}[2]{\setlength{\tabcolsep}{#1}\renewcommand{\arraystretch}{#2}\centering\footnotesize}

\title{Learning and Reasoning with Geometry: How Generation Helps Spatial Understanding}

\author[1]{Rui Yang}
\author[2,\emailicon]{Yanwei Li}
\author[3]{Sili Chen}
\author[3]{Jingjia Huang}
\author[3]{Yi Lin}
\author[1,\emailicon]{Hengshuang Zhao}

\affiliation[1]{The University of Hong Kong}
\affiliation[2]{Shanghai Jiao Tong University}
\affiliation[3]{ByteDance Seed}

\abstract{

Image and video generators can produce spatially consistent visual content, but how to use this capability to improve spatial understanding remains underexplored.
We study this problem from two perspectives: geometric supervision during training and explicit geometric evidence during inference.
To this end, we introduce a unified geometric understanding and generation model (GeoUG) that integrates depth and multi-view generation with spatial understanding.
First, we design a multi-scale depth codec that enables metric-depth generation using a pretrained visual autoencoder, and show that joint training yields task-dependent gains in spatial understanding.
Second, we introduce 3DCueBench to control geometric difficulty while balancing selected 2D cues, and find that generated depth provides larger reasoning gains on questions requiring fine-grained geometric discrimination.
Building on this finding, we develop adaptive interleaved reasoning, in which GeoUG first drafts an answer and then decides whether to generate geometric evidence to confirm or revise it.
We train this decision through reinforcement learning, balancing accuracy gains over retaining the draft against generation cost.
GeoUG achieves circular accuracies of 97.9\%, 90.8\%, and 94.2\% on CV-Bench-3D, 3DCueBench, and DA-2K with fewer generation calls than always generating.

}

\date{\today}
\checkdata[Code]{\url{https://github.com/hku-sail/GeoUG}}

\begin{document}
\maketitle
\etocdepthtag.toc{mtmain}


\section{Introduction}
\label{sec:introduction}

Foundation models increasingly learn from multimodal observations of the world, extending beyond the knowledge conveyed through language~\citep{rw_ijepa,rw_vjepa2,rw_beyondlm}.
While language efficiently expresses concepts and relations, it often leaves the fine-grained geometry of physical scenes and their continuous changes underspecified~\citep{rw_talmy_space}.
Learning to predict visual observations offers a complementary source of supervision, requiring models to capture spatial structure and how it changes across views and over time.
Recent advances in visual generation and editing~\citep{rw_seedance2,rw_minimax_h3,rw_gemini_flash_image} suggest growing capabilities to maintain coherent scene structure through such changes.
Recent studies of multimodal pretraining further highlight the potential for generation and understanding to benefit from shared learning, while revealing that knowledge transfer is asymmetric and task-dependent~\citep{rw_beyondlm,rw_physics}.
This raises a concrete question for geometry, which underlies both coherent visual generation and accurate spatial judgments: \textit{how can geometric generation help models learn spatial relations and reason about them?}

To address this question, we examine geometric generation during both learning and reasoning.
During learning, visual question answering (VQA) typically supervises the specific relations queried in a scene, whereas generating depth or images from other viewpoints provides supervision over richer scene structure and its visual consequences.
This motivates examining how different geometric generation objectives contribute to spatial understanding.
During reasoning, generated geometry can provide explicit intermediate representations from which depth and distance measurements can be obtained to support spatial judgments.
We therefore ask when using this additional evidence improves an answer and whether models can learn to acquire it selectively.
These two perspectives connect learning spatial relations from geometric supervision with reasoning over generated geometry.

\begin{figure}[t]
  \centering
  \includegraphics[width=\textwidth]{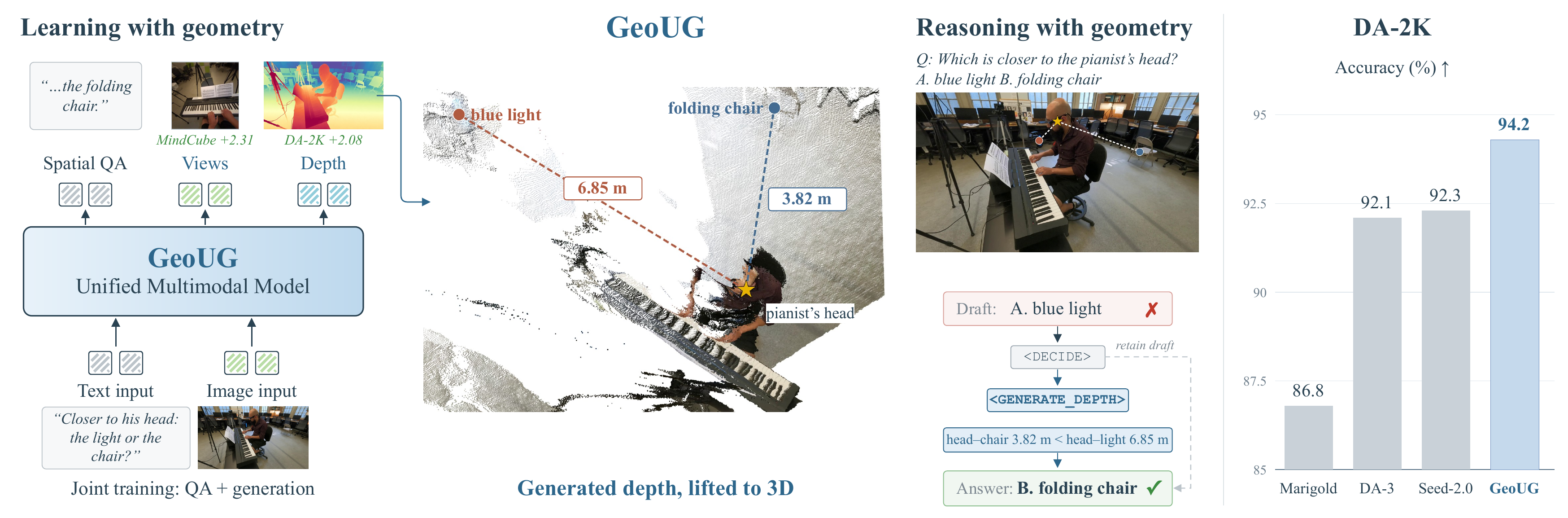}
  \caption{\textbf{Learning and reasoning with geometry.} \emph{Learning with geometry} (left): GeoUG co-trains spatial understanding with geometric generation of new views and metric depth, improving MindCube and DA-2K (Section~\ref{sec:cotraining}). \emph{Reasoning with geometry} (middle and right): GeoUG first drafts an answer, then decides whether to generate depth. Here it generates depth, queries 3D distances from the generated depth lifted to 3D, and revises the incorrect draft. Far right: comparison of GeoUG with Marigold~\citep{ke2024marigold}, DA3~\citep{da3}, and Seed-2.0~\citep{seed2} on DA-2K~\citep{depthanythingv2}.}
  \label{fig:teaser}
\end{figure}

To study these two roles, we develop GeoUG (Figure~\ref{fig:teaser}), a unified model that integrates spatial understanding with metric-depth generation and view-generation tasks.
To enable metric-depth generation, we introduce a Multi-scale Metric Depth Codec (MMDC).
MMDC encodes metric depth into a bounded three-channel representation compatible with a pretrained visual autoencoder, with each channel operating at a different depth scale.
During training, we jointly optimize a next-token prediction objective for spatial understanding and a flow-matching objective for geometric generation.
Comparing different geometric generation objectives reveals task-dependent benefits: metric-depth generation primarily improves depth-related spatial abilities, whereas view generation benefits spatial understanding across viewpoints.
During inference, GeoUG uses generated metric depth as explicit geometric evidence.
To characterize when this evidence is useful, we introduce 3DCueBench, which balances cases in which selected 2D cues agree or conflict with the true 3D relations across different geometric difficulty levels.
We find that generated depth provides larger gains when the compared objects have smaller depth or distance gaps.
Motivated by this finding, we develop adaptive interleaved reasoning, in which GeoUG first drafts an answer and then decides whether to generate geometric evidence to confirm or revise it.
We train this draft-conditioned policy using reinforcement learning with verifiable rewards (RLVR)~\citep{rw_deepseek_r1}, introducing a decision-specific advantage that compares generation with retaining the draft while accounting for generation cost.
By acquiring geometric evidence selectively, GeoUG retains most of the gains from always generating while requiring fewer generation calls.
It achieves circular accuracies of 97.9\%, 90.8\%, and 94.2\% on CV-Bench-3D~\citep{rw_cambrian}, 3DCueBench, and DA-2K~\citep{depthanythingv2}, respectively.
Together, these findings highlight the task-dependent benefits of geometric supervision during training and the value of selectively acquiring geometric evidence during inference.

Our main contributions are as follows:
\begin{itemize}[leftmargin=1.5em,topsep=2pt,itemsep=2pt,parsep=0pt]
    \item We study geometric generation as both training supervision and reasoning evidence, analyzing the task-dependent benefits of different generation objectives and how inference gains vary with geometric difficulty.
    \item We introduce GeoUG, which integrates metric-depth and camera-conditioned multi-view generation into a unified understanding and generation model. Using the draft answer as a reference, reinforcement learning balances accuracy gains against generation costs to support on-demand generation and use of geometric evidence.
    \item We introduce 3DCueBench, which controls geometric difficulty while balancing the validity of selected 2D cues, enabling a more targeted evaluation of spatial reasoning. Across three spatial understanding benchmarks, GeoUG achieves the highest circular accuracy among the evaluated vision-language models, while using fewer generation calls and maintaining accuracy close to or above the always-generate configuration.
\end{itemize}

\section{Related Work}
\label{sec:related_work}

\noindent \textbf{Unified Understanding and Generation.}
Unified multimodal models combine language prediction with visual generation~\citep{rw_showo,rw_transfusion}.
BAGEL uses a Mixture-of-Transformer-Experts (MoT) architecture with specialized understanding and generation experts~\citep{bagel}.
MetaMorph reports mutual benefits with asymmetric contributions from understanding and generation data~\citep{rw_metamorph}.
Recent pretraining studies investigate cross-modal complementarity, knowledge transfer, and conditions for synergy~\citep{rw_beyondlm,rw_physics}.
We examine these interactions specifically through geometric generation and spatial understanding.

\noindent \textbf{Generation for Understanding and Reasoning.}
Generative supervision improves visual understanding through reconstruction~\citep{rw_ross} and semantic proxy tasks~\citep{rw_sgt}.
At inference time, MVoT generates visual reasoning traces~\citep{rw_mvot}, while visual world-modeling studies identify task-dependent benefits of generated intermediates~\citep{rw_visworld}.
COOPER learns relative-depth and segmentation generation alongside adaptive interleaved spatial reasoning~\citep{rw_cooper}.
For cross-view reasoning, recent work examines synthesized viewpoints and whether models effectively use generated visual evidence~\citep{rw_novelviews,rw_imagine}.
Our work develops multi-scale metric-depth encoding and studies depth and view generation as training supervision and generated depth as evidence during inference, emphasizing how geometric difficulty and generation errors affect downstream benefits.
This analysis distinguishes the benefits of learning geometric representations from those of explicitly generating and interpreting them at test time.

\noindent \textbf{Spatial Understanding and Reasoning.}
Existing approaches strengthen spatial capabilities through geometric representations and spatial supervision.
VLM-3R incorporates a geometry encoder to extract implicit 3D tokens from monocular video~\citep{rw_vlm3r}.
From a data perspective, SpatialVLM, Visual Spatial Tuning, and OpenSpatial construct spatial instruction data to improve perception and reasoning~\citep{rw_spatialvlm,vst,rw_openspatial}.
Their instruction targets provide spatially sparse supervision through selected attributes, objects, and relations, rather than dense per-pixel geometric targets.
We complement such supervision with metric-depth and multi-view generation objectives, providing dense visual supervision for learning geometry and enabling generated representations to support spatial reasoning.

\section{Method}

\label{sec:method}

We investigate how geometric generation helps spatial understanding through
learning and reasoning with geometry. We build GeoUG on BAGEL~\citep{bagel},
a unified understanding and generation model. For learning with geometry,
we study co-training configurations that jointly optimize spatial understanding
and a geometric generation task, using cross-entropy loss on text tokens and
flow-matching loss for visual generation. For reasoning with geometry, the
model first drafts an answer and then decides whether to generate depth and
query geometric measurements to confirm or revise it. We train this decision
to balance answer correctness against generation cost.

\subsection{Geometric Generation Tasks}
\label{sec:geometric-generation}
\label{sec:mmdc}

We unify geometric generation tasks through a common visual output
representation and velocity-prediction objective based on flow
matching~\citep{lipman2023flow,liu2023flow}. Each target is represented
as a three-channel image: an encoded metric-depth map or an RGB target view.
Following BAGEL, a frozen VAE encodes this target into latents, and the model
predicts the flow velocity from noise-corrupted latents conditioned on the
input image, task instruction, and timestep. At inference, the predicted
velocity field transforms noise into target latents, which the VAE decodes
into the visual output. The tasks thus share one generation mechanism and differ only in conditioning and target.

We study two forms of geometric supervision within this framework.
\textbf{Metric-depth generation} captures the depths of visible surfaces,
providing explicit geometry for spatial measurement.
\textbf{View generation} captures changes in spatial layout,
visibility, and occlusion induced by a change of viewpoint. In
camera-conditioned multi-view generation (\emph{multi-view} for short),
the target viewpoint is given by a camera-motion
instruction, which we derive from relative camera poses to account for both
rotation and translation (Appendix~\ref{app:multiview-data}). In
exo-to-ego generation, the target viewpoint is that of a person in a
third-person image; generating that person's first-person view requires
perspective taking, imagining the scene from another observer's viewpoint
(Appendix~\ref{app:exo-ego-data}).

\noindent\textbf{Multi-scale Metric Depth Codec.}
To represent metric depth within this common image interface, we introduce
the Multi-scale Metric Depth Codec (MMDC). It maps depth into a bounded
three-channel representation compatible with the pretrained VAE and converts
the generated output back into metric depth, preserving a consistent physical
scale across images.

\begin{figure}[t]
  \centering
  \begin{minipage}[c]{0.58\linewidth}
    \centering
    \includegraphics[width=\linewidth]{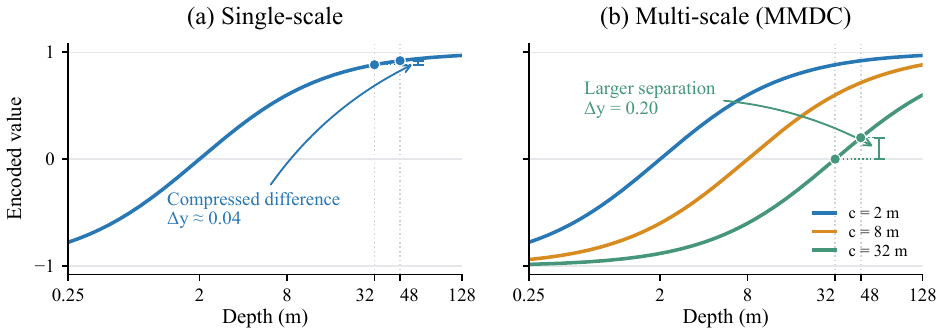}
  \end{minipage}\hfill
  \begin{minipage}[c]{0.39\linewidth}
    \caption{\textbf{Multi-scale depth encoding.} (a) A single pivot compresses depth differences far from its scale. (b) A second channel with a larger pivot separates depths saturated in the first.}
    \label{fig:mmdc-insight}
  \end{minipage}
\end{figure}

Generative depth estimators such as Marigold~\citep{ke2024marigold}
normalize each ground-truth depth map using its scene-specific 2nd and 98th
percentiles. Given a metric ground-truth depth value $d$ at a pixel, they
construct the normalized target
$\tilde d=2\left((d-d_2)/(d_{98}-d_2)-1/2\right)$, where $d_2$ and $d_{98}$
are computed independently for each depth map.
This produces an affine-invariant target compatible with the bounded input
range of an RGB VAE. Because the metric values of $d_2$ and $d_{98}$ are
unavailable at inference, the predicted normalized depth cannot be converted
back to metric depth without recovering the unknown scene-specific scale and
shift.
We instead encode the same metric depth $d>0$ as a normalized value
$y\in(-1,1)$ using a fixed, invertible transformation with a fixed pivot
$c>0$, where $d$ and $c$ are both in metres:
\begin{equation}
  y=\frac{d-c}{d+c}\in(-1,1),\qquad
  d=c\frac{1+y}{1-y}.
  \label{eq:mmdc-single}
\end{equation}
Every encoded value now has a consistent metric meaning across images,
allowing the generated representation to be decoded into a metric depth map
at inference.
However, a single pivot $c$ compresses relative depth differences away from
its characteristic scale, amplifying reconstruction errors during decoding,
as shown in Figure~\ref{fig:mmdc-insight}(a).
MMDC addresses this by exploiting the three input channels of the RGB VAE.
Rather than repeat a
single-scale encoding three times, we assign each channel a different depth
scale:
\begin{equation}
  Y_i=\frac{d-c_i}{d+c_i},\qquad
  (c_1,c_2,c_3)=(2,8,32)\,\mathrm{m}.
  \label{eq:encode-multi}
\end{equation}
The channels provide complementary precision across depth ranges: where one
channel approaches saturation, another can better distinguish depth
differences (Figure~\ref{fig:mmdc-insight}). The pivots are fixed across
datasets, and the resulting representation fits the pretrained VAE without
architecture changes.

For decoding, we invert each generated channel and fuse the per-channel
estimates:
\begin{equation}
  \hat d_i=c_i\frac{1+\hat Y_i}{1-\hat Y_i},\qquad
  w_i=(1-\hat Y_i^2)^2,\qquad
  \hat d=\frac{\sum_i w_i\hat d_i}{\sum_i w_i}.
  \label{eq:decode}
\end{equation}
The weight is largest near a channel's pivot and decreases towards saturation,
allowing different channels to contribute at different depth scales.
More details are provided in Appendix~\ref{app:mmdc}.

MMDC introduces no learned parameters. During training, encoded depth maps
serve as targets through the frozen VAE encoder. At inference, MMDC converts
the VAE-decoded outputs into metric depth maps for geometric measurement and
interleaved reasoning.

\subsection{Interleaved Geometric Reasoning}
\label{sec:interleaved-reasoning}

We next turn from geometric supervision during training to using generated
depth as evidence for spatial reasoning at inference time. Before
designing the reasoning procedure, we address two questions through two pilot
studies:

\begin{figure}[t]
  \centering
  \begin{minipage}[t]{0.4818\linewidth}
    \vspace{0pt}
    \centering
    \includegraphics[width=\linewidth]{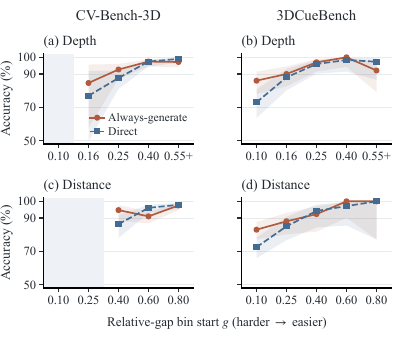}
    \caption{\textbf{When does generation help?}
    Accuracy versus relative gap $g$; smaller is harder.
    Non-circular accuracy from separately trained models.
    Generation gives larger gains on difficult 3DCueBench questions.
    Gains on easier questions are limited.}
    \label{fig:difficulty-by-task}
  \end{minipage}\hfill
  \begin{minipage}[t]{0.4764\linewidth}
    \vspace{0pt}
    \centering
    \includegraphics[width=\linewidth]{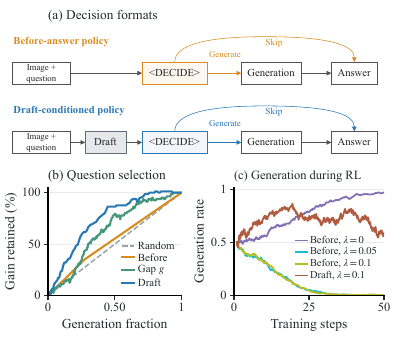}
    \caption{\textbf{When to generate.}
    (a) Policies decide whether to generate before answering or after a draft.
    (b) Gain retained at different generation budgets on 3DCueBench
    (protocol in Appendix~\ref{sec:app-adaptive}).
    Gap $g$ prioritizes questions with smaller ground-truth relative gaps.
    (c) Generation rates during RL with generation cost $\lambda$.}
    \label{fig:rl-pilot}
  \end{minipage}
\end{figure}

\noindent\textbf{Pilot I: When Does Generation Help?}
\label{sec:difficulty}
We compare direct answering (\emph{Direct}) with reasoning that always generates depth before answering (\emph{Always-generate}) on CV-Bench-3D~\citep{rw_cambrian} and 3DCueBench
(Section~\ref{sec:eval-d3}). We measure difficulty by the relative gap
\begin{equation}
  g=\frac{|d_a-d_b|}{\max(d_a,d_b)},
  \label{eq:relative-gap}
\end{equation}
where $d_a$ and $d_b$ are the camera depths or distances to the anchor
being compared. We group questions in both datasets into bins according to
their relative gap. Figure~\ref{fig:difficulty-by-task} shows larger gains
from generation-assisted reasoning over direct answering on difficult
questions with smaller gaps. On easier questions, direct answering already
achieves high accuracy, while generation-assisted reasoning provides limited
gains and sometimes reduces accuracy.
This degradation may result from errors in generated depth or geometric
measurements. These findings motivate adaptive interleaved reasoning,
in which the model decides when to generate geometric evidence.

\noindent\textbf{Pilot II: Can the Model Tell When It Needs Geometry?}
\label{sec:pilot-when}
The ground-truth gap is unavailable at test time, so the model must decide
whether to generate from its own reasoning context. A straightforward
approach is a \emph{before-answer policy}, which chooses between direct
answering and generation-assisted reasoning before producing an answer
(Figure~\ref{fig:rl-pilot}a, top). To evaluate whether the policy identifies
questions that benefit from generation, we use its \emph{generation probability}:
the probability of choosing the depth-generation branch at the
\texttt{<DECIDE>} step in Figure~\ref{fig:rl-pilot}a. For each generation
budget, questions with the highest generation probabilities receive
generation-assisted reasoning, while the remainder are answered directly,
yielding the orange \emph{Before} curve in Figure~\ref{fig:rl-pilot}b. In contrast,
the green \emph{Gap $g$} curve ranks questions by their ground-truth relative
gaps in ascending order, prioritizing the smallest-gap questions.
The vertical axis reports the fraction of the accuracy gain from always
generating retained by each selection strategy on 3DCueBench
(the detailed protocol can be found in Appendix~\ref{sec:app-adaptive}).
The Gap $g$ reference requires ground-truth geometry and is
unavailable at test time. Random selection retains a fraction of the gain
equal to the generation budget in expectation.

At a 50\% generation budget, the before-answer policy retains only
55\% of the gain from always generating, close to the 50\% expected from
random selection (Figure~\ref{fig:rl-pilot}b). Reinforcement learning also
fails to produce selective generation in our runs: the policy tends to
generate for either almost no questions or almost all questions
(Figure~\ref{fig:rl-pilot}c). These results suggest that, in this setup, the
before-answer policy does not learn to identify questions that benefit from
generation. We draw inspiration from reflective reasoning in chain-of-thought,
where a model forms a tentative answer and subsequently reconsiders it before
reaching a final answer. This motivates us to use the tentative answer as a
draft and place the generation decision at this point of reconsideration.

Specifically, we introduce a \emph{draft-conditioned policy}, which first
produces an initial draft from the input image and question without depth
generation, then decides whether to retain it as the final answer or generate
geometric evidence to verify or revise it
(Figure~\ref{fig:rl-pilot}a, bottom).
Under the same ranking protocol and a 50\% generation budget, this policy
retains 86\% of the gain from always generating, compared with 55\% for the
before-answer policy (Figure~\ref{fig:rl-pilot}b).
Each policy's retained gain is normalized by its own direct and
always-generate accuracies.
We instantiate this policy with depth generation. For example, to decide
which of two objects is closer, the model generates a depth map, decodes it
into metric depth using MMDC, and queries the objects' depths before
confirming or revising its draft.

\noindent\textbf{Learning When to Generate.}
\label{sec:adaptive-reasoning}
We train the draft-conditioned policy in two stages. First, we cold-start
the policy through supervised fine-tuning on chain-of-thought (CoT) sequences
that follow the draft-conditioned reasoning format: an initial answer draft,
a decision on whether to generate depth, and a final answer that either
retains the draft or uses generated geometric evidence to verify or revise it.
Draft tokens are excluded from the training loss
(Appendix~\ref{sec:app-adaptive}).

Starting from this initialization, we use GSPO-based reinforcement
learning~\citep{gspo} to optimize the policy's generation decisions for the
accuracy--cost trade-off. The reward is $r_i=c_i-\lambda a_i$, where
$c_i\in\{0,1\}$ indicates final-answer correctness, $a_i\in\{0,1\}$ indicates
generation, and $\lambda$ is the generation cost.

For $G$ rollouts of a question, GSPO assigns each a shared trajectory-level advantage:
\begin{equation}
  \bar r=\frac{1}{G}\sum_{j=1}^{G}r_j,
  \qquad
  A_i^{\mathrm{group}}=\frac{r_i-\bar r}{s_r+\epsilon},
  \label{eq:group-adv}
\end{equation}
where $s_r$ is the standard deviation of rewards within the group and
$\epsilon>0$ ensures numerical stability. However, when every rollout
generates, the generation cost cancels:
\begin{equation}
  r_i-\bar r=(c_i-\lambda)-(\bar c-\lambda)=c_i-\bar c,
  \qquad
  \bar c=\frac{1}{G}\sum_{j=1}^{G}c_j.
  \label{eq:generation-cost-cancellation}
\end{equation}
Since subtracting a shared cost also leaves the standard deviation unchanged,
the advantage no longer depends on $\lambda$. In such groups, the decision
receives no cost-dependent signal for learning whether generation is
worthwhile. In practice we replace $s_r$ with a constant scale, since normalizing by $s_r$
also removes the magnitude of $\lambda$ (Appendix~\ref{sec:app-adaptive});
the cancellation above holds for either choice.

We therefore assign the generation decision an advantage that explicitly
compares generation with retaining the draft. Because answering directly
returns the draft unchanged, its correctness $c_{d,i}$ is known from the answer label, even for rollouts that generate. For a sampled
decision, we use
\begin{equation}
  A_i^{\mathrm{gate}} =
  \begin{cases}
    (1-p_i)\,(c_i-\lambda-c_{d,i}) & a_i=1,\\[3pt]
    -p_i\,(\widehat Q_{\mathrm{gen},i}-\lambda-c_{d,i}) & a_i=0,
  \end{cases}
  \label{eq:gate-adv}
\end{equation}
where $p_i$ is the generation probability conditioned on the draft, and
$\widehat Q_{\mathrm{gen},i}$ estimates correctness after generation using
forced-generation rollouts for the same question. This preserves a signal about the benefit and cost of generation even when all sampled rollouts choose the same branch.

\section{Evaluation}
\label{sec:evaluation}

\subsection{3DCueBench: Controlled Evaluation of Geometric Reasoning}
\label{sec:eval-d3}

\begin{figure}[t]
  \centering
  \includegraphics[width=\linewidth]{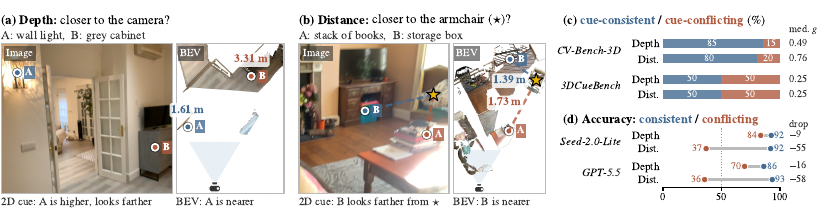}
  \caption{\textbf{2D shortcuts versus 3D geometry.} (a) ``Lower is closer'' picks the cabinet, but the bird's-eye view (BEV) shows the higher wall light is nearer. (b) The books are closer to the armchair in the image plane, but the storage box is closer in 3D. (c) Share of cue-consistent (2D rule correct) and cue-conflicting (2D rule wrong) questions, with median relative gap $g$. (d) Ordinary accuracy of two VLMs on cue-consistent vs.\ cue-conflicting 3DCueBench questions (Appendix Table~\ref{tab:d3-results}).}
  \label{fig:3dcuebench-examples}
\end{figure}

VLMs may rely on 2D image-plane relationships to answer 3D spatial questions
without understanding the underlying 3D structure.
Figure~\ref{fig:3dcuebench-examples}(a,b) shows two such shortcuts,
``lower in the image is closer'' and ``closer in the image plane is closer
in 3D'', each contradicted by the scene's 3D geometry.
In our analysis of CV-Bench-3D, choosing the object with the lower
bounding-box bottom correctly answers 85.3\% of Depth questions
(Figure~\ref{fig:3dcuebench-examples}(c)), and no question has a relative gap below 0.16.
Current VLMs follow such cues: on 3DCueBench Distance questions where the rule
gives the wrong answer, Seed-2.0-Lite~\citep{seed2} and GPT-5.5~\citep{gpt55} reach only
37.0\% and 35.7\% (Figure~\ref{fig:3dcuebench-examples}(d)).
This makes it difficult to determine whether high accuracy reflects geometric
reasoning or reliance on image-plane cues.

We therefore introduce \textbf{3DCueBench}, a benchmark of 800 carefully
selected questions in two tasks: Depth compares the camera depths of two
targets, and Distance compares their 3D distances to an anchor. In half of
the questions (\emph{cue-consistent}), the 2D rules used to construct the
benchmark favor the correct answer; in the other half
(\emph{cue-conflicting}), they favor the wrong answer. Each rule therefore
achieves 50\% accuracy, so models must go beyond these rules to perform well.
To control difficulty, we sample questions across intervals of the relative gap (Equation~\ref{eq:relative-gap}), keeping cue-consistent and cue-conflicting questions balanced within each interval (Appendix~\ref{app:d3}).

\subsection{Evaluation Protocol}
\label{sec:eval-protocol}

\noindent\textbf{Benchmarks and Metrics.}
We evaluate single-image depth and distance reasoning on CV-Bench-3D~\citep{rw_cambrian},
DA-2K~\citep{depthanythingv2}, and our 3DCueBench.
Unless suffixed \emph{-Visual}, CV-Bench-3D and DA-2K denote their textual versions,
which give object points or boxes as coordinates in the question text.
We further assess multi-view spatial reasoning on MindCube~\citep{wang2025mindcube}
and the camera-motion subset of MMSI-Bench~\citep{yang2025mmsibench},
with accuracy or circular accuracy as noted in each figure or table.
For dense depth estimation, we evaluate on NYU-v2~\citep{nyuv2},
KITTI~\citep{kitti}, ETH3D~\citep{eth3d}, Sintel~\citep{sintel}, and
DIODE~\citep{diode}, reporting AbsRel and $\delta_1$ (threshold $1.25$).

\section{Experiment}

\subsection{Generative Co-training}
\label{sec:cotraining}

\begin{figure}[htbp]
  \centering
  \includegraphics[width=\textwidth]{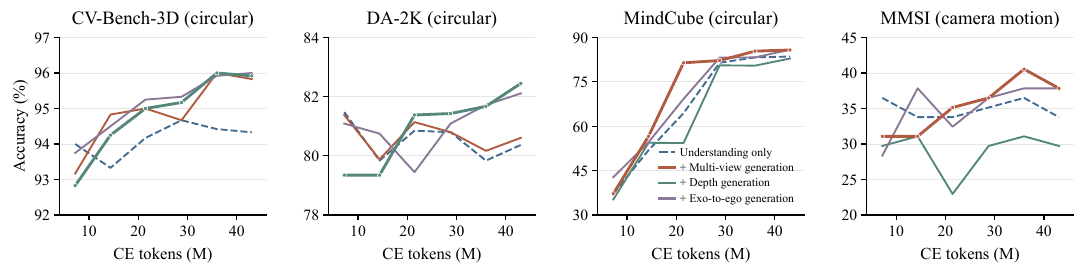}
  \caption{\textbf{Generative co-training.} Circular accuracy on CV-Bench-3D-Visual, DA-2K-Visual, and MindCube; accuracy on MMSI camera motion. x-axis: understanding CE tokens.}
  \label{fig:cotraining}
\end{figure}

\noindent\textbf{Generative co-training yields task-dependent gains in spatial understanding.}
We compare understanding-only training with three co-training configurations, each combining spatial understanding with one geometric generation task introduced in Section~\ref{sec:geometric-generation}: metric-depth generation, camera-conditioned multi-view generation (multi-view), or exo-to-ego generation (exo-to-ego). Figure~\ref{fig:cotraining} plots benchmark accuracy against the number of understanding CE tokens consumed during training. All three tasks improve final accuracy on CV-Bench-3D-Visual by 1.50--1.67 points. Depth generation gives the largest gain on DA-2K-Visual (+2.08 points), while multi-view and exo-to-ego generation improve MindCube by 2.21 and 2.31 points. On MMSI camera motion, multi-view generation improves accuracy, whereas depth generation lowers it. These results suggest that the benefits depend on how closely the generation task matches the spatial ability being evaluated. Appendix~\ref{app:gen-benchmark} reports the generation quality of the co-trained models, and Appendix~\ref{app:pretraining-sft} shows that generation pretraining followed by understanding-only SFT does not reliably yield such gains.

\begin{figure}[t]
  \centering
  \scriptsize
  \setlength{\tabcolsep}{0pt}
  \renewcommand{\arraystretch}{1}
  \begin{subfigure}{\linewidth}
    \centering
    \begin{tabular}{c@{\hspace{2pt}}c@{\hspace{2pt}}c@{\hspace{10pt}}c@{\hspace{2pt}}c@{\hspace{2pt}}c}
      Input & Generated depth & 3D reconstruction & Input & Generated depth & 3D reconstruction \\[1pt]
      \genimg{48pt}{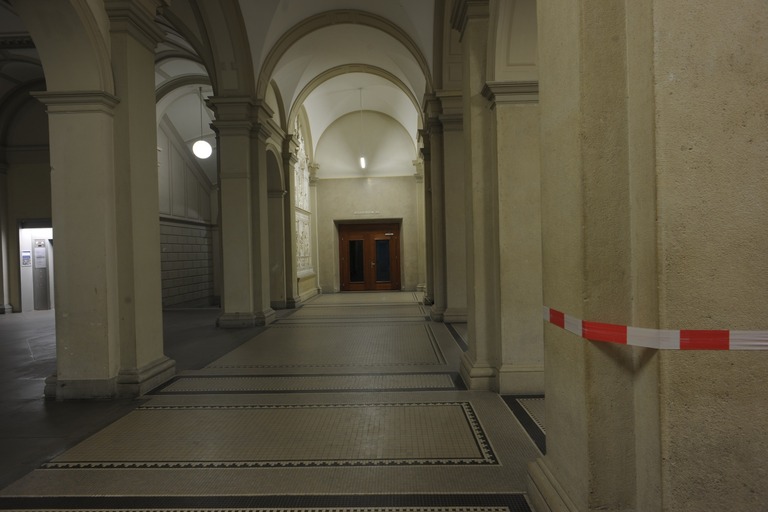} & \genimg{48pt}{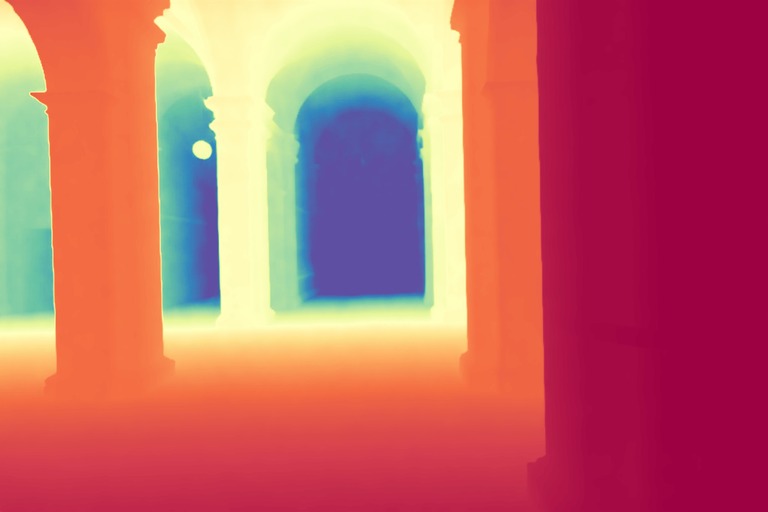} & \genimg{48pt}{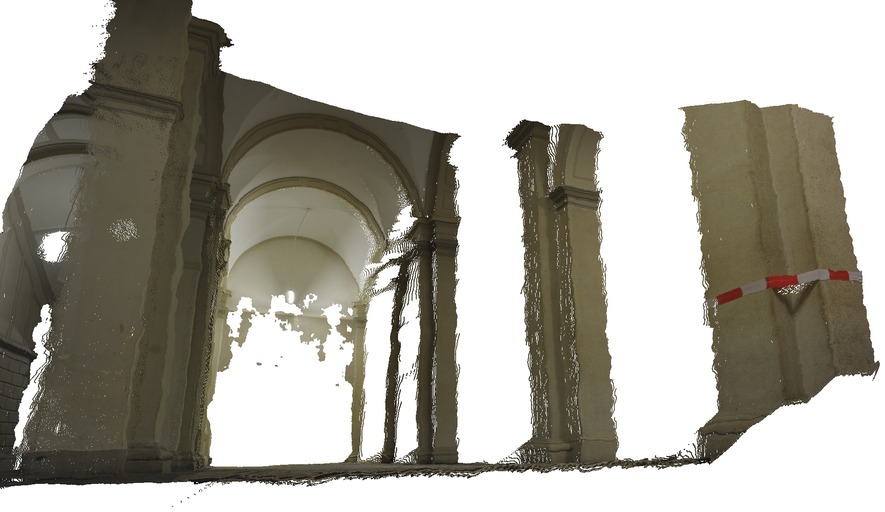} &
      \genimg{48pt}{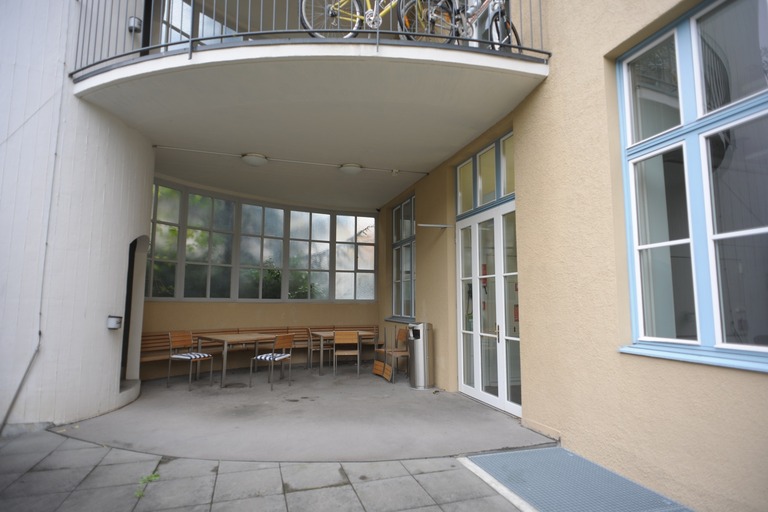} & \genimg{48pt}{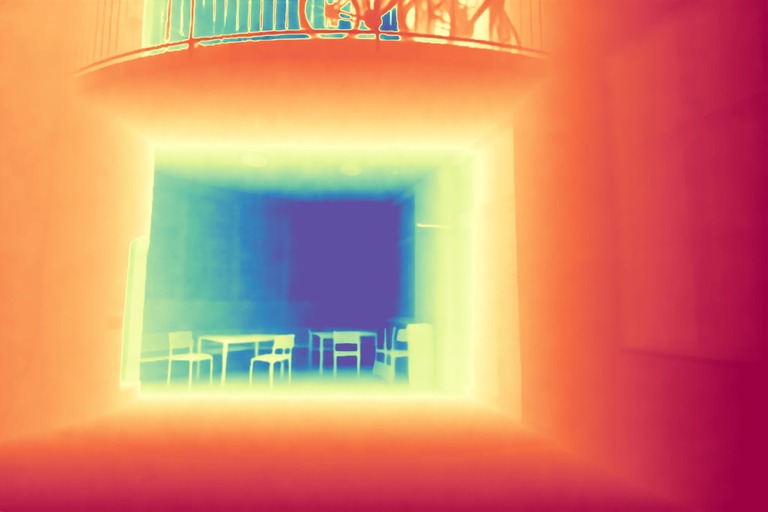} & \genimg{48pt}{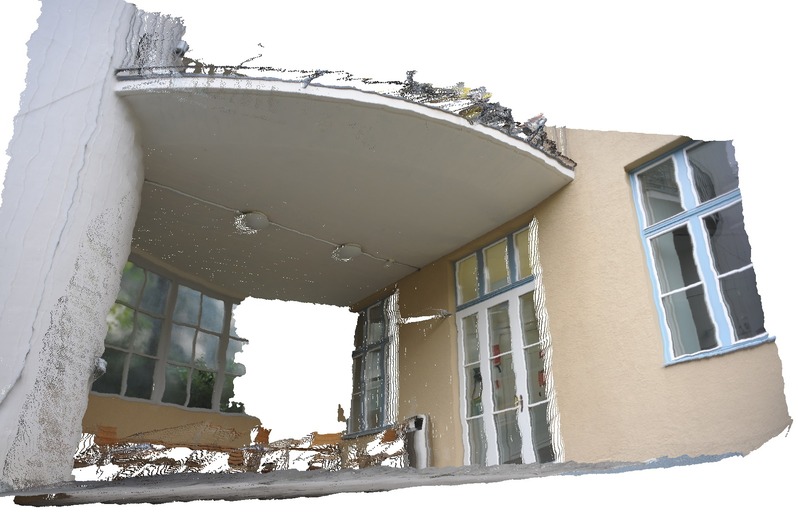} \\
    \end{tabular}
    \caption{\textbf{Depth generation} on ETH3D~\citep{eth3d}: metric depth and its back-projected 3D reconstruction.}
    \label{fig:gen-main-depth}
  \end{subfigure}\\[4pt]
  \begin{subfigure}{\linewidth}
    \centering
    \begin{tabular}{c@{\hspace{2pt}}c@{\hspace{2pt}}c@{\hspace{10pt}}c@{\hspace{2pt}}c@{\hspace{2pt}}c}
      Input & BAGEL & Ours & Input & BAGEL & Ours \\[1pt]
      \genimg{42pt}{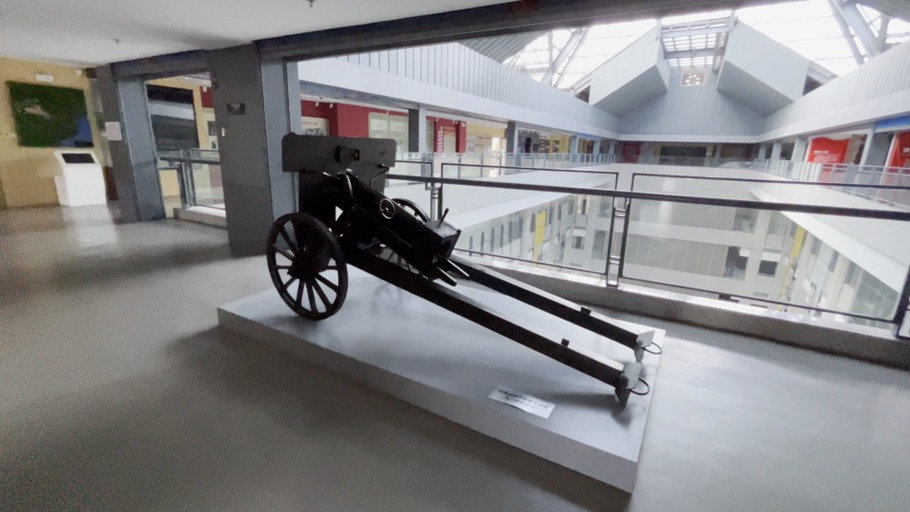} & \genimg{42pt}{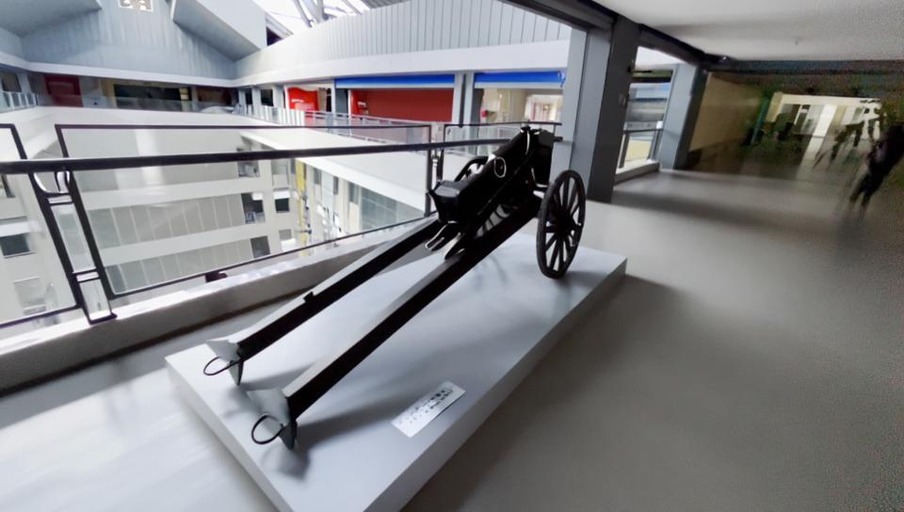} & \genimg{42pt}{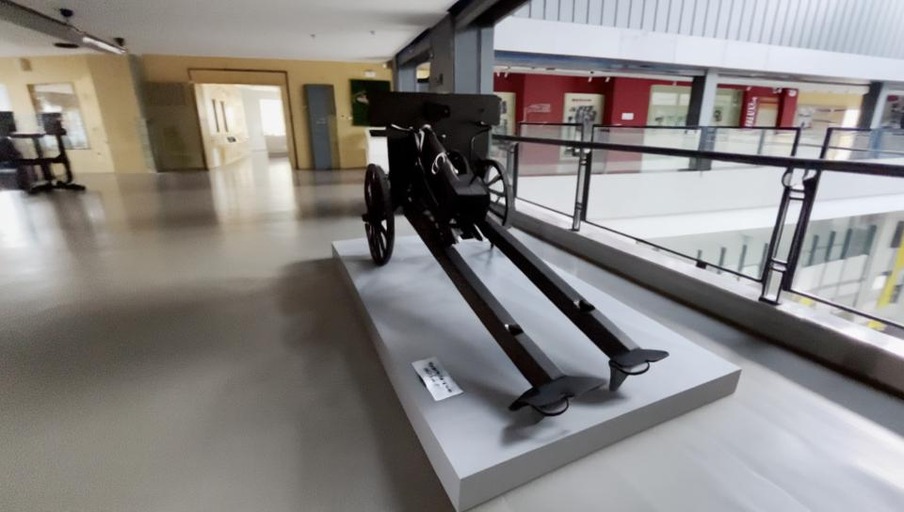} &
      \genimg{42pt}{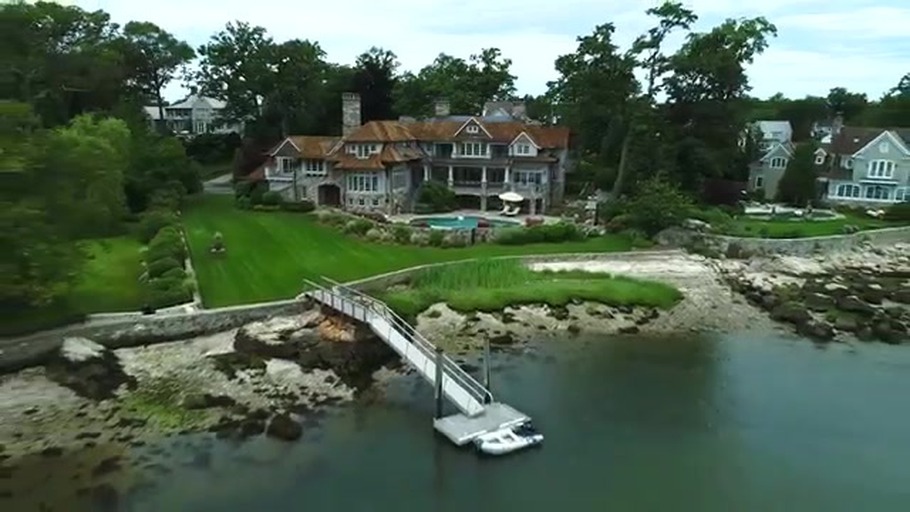} & \genimg{42pt}{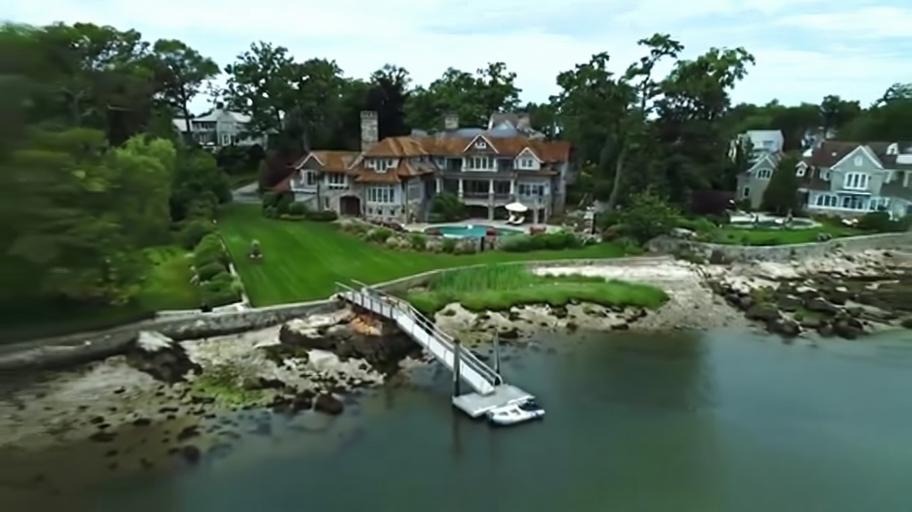} & \genimg{42pt}{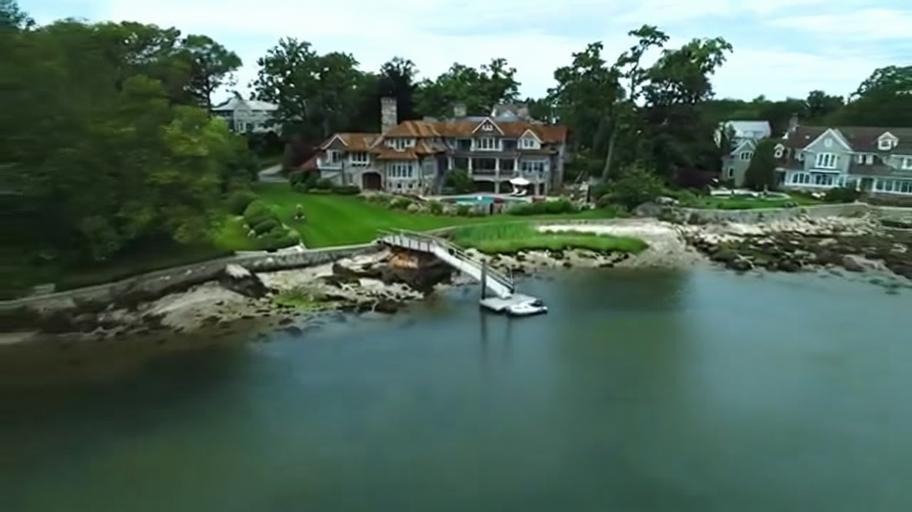} \\[1pt]
      \multicolumn{3}{p{228pt}}{\raggedright\itshape Move the camera right 0.88\,m and rotate the camera left 20 degrees} &
      \multicolumn{3}{p{228pt}}{\raggedright\itshape Move the camera backward 1.29\,m} \\
    \end{tabular}
    \caption{\textbf{Multi-view generation} on DL3DV~\citep{dl3dv} (left) and RealEstate10K~\citep{re10k} (right).}
    \label{fig:gen-main-mv}
  \end{subfigure}\\[4pt]
  \begin{subfigure}{\linewidth}
    \centering
    \begin{tabular}{c@{\hspace{2pt}}c@{\hspace{2pt}}c@{\hspace{10pt}}c@{\hspace{2pt}}c@{\hspace{2pt}}c}
      Input & BAGEL & Ours & Input & BAGEL & Ours \\[1pt]
      \genimg{58pt}{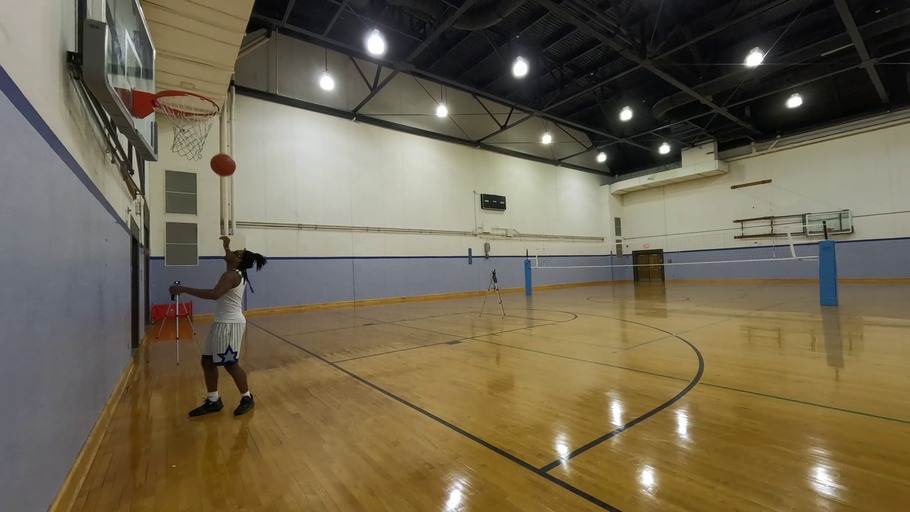} & \genimg{58pt}{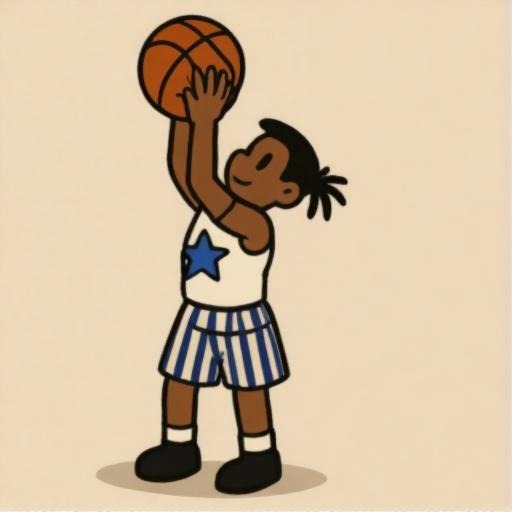} & \genimg{58pt}{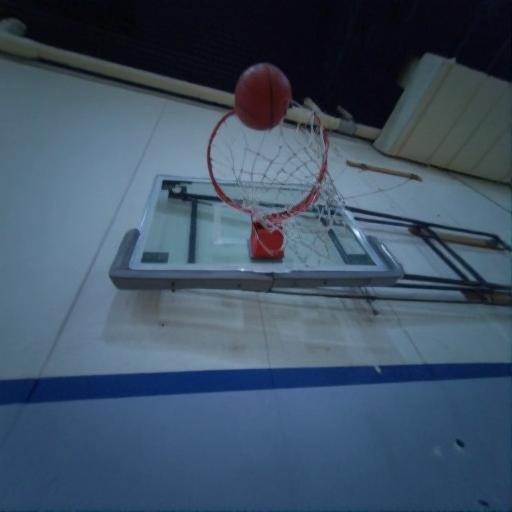} &
      \genimg{58pt}{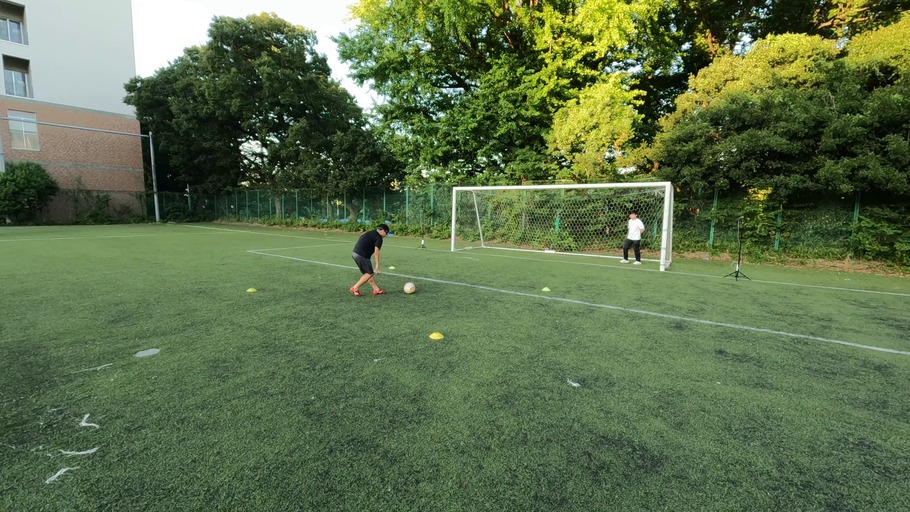} & \genimg{58pt}{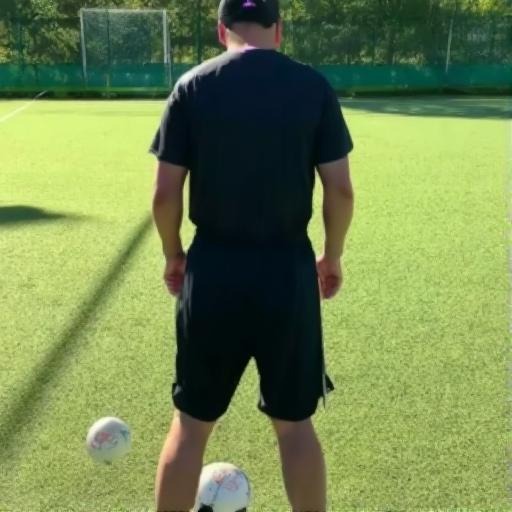} & \genimg{58pt}{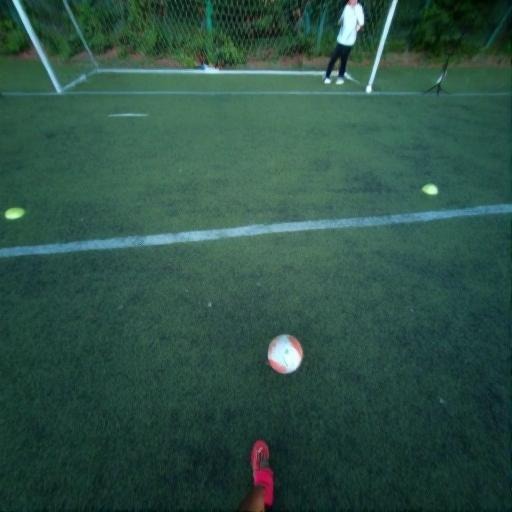} \\[1pt]
      \multicolumn{3}{p{223pt}}{\raggedright\itshape Show me the ego view of the person performing a Mikan layup basketball drill (wearing a white tank top, striped shorts with a blue star, and black shoes)} &
      \multicolumn{3}{p{223pt}}{\raggedright\itshape Show me the ego view of the person in black clothing preparing to take a soccer penalty kick.} \\
    \end{tabular}
    \caption{\textbf{Exo-to-ego generation} on Ego-Exo4D~\citep{grauman2024ego}.}
    \label{fig:gen-main-exo-ego}
  \end{subfigure}
  \caption{\textbf{Geometric generation after co-training.} BAGEL~\citep{bagel} is the original model; Ours are the multi-view and exo-to-ego co-training models (Figure~\ref{fig:cotraining}) at their final checkpoints. Instructions are shown below each example.}
  \label{fig:gen-main}
\end{figure}

\noindent\textbf{Co-training also teaches geometrically controlled generation.}
Figure~\ref{fig:gen-main} shows the three geometric generation abilities. For depth generation, the generated metric depth preserves the scene layout, such as the receding arches and free-standing pillars of the hallway and the overhanging roof of the terrace, and back-projects to coherent 3D reconstructions. For multi-view generation, the original BAGEL~\citep{bagel} often returns a view close to the input, whereas our model moves the camera as instructed and brings new content into view: it swings the view around the cannon when moving right and turning left, and pulls back to reveal the shoreline when moving backward, while BAGEL nearly copies the input. For exo-to-ego generation, BAGEL depicts the described person, whereas our model renders the court from that person's viewpoint, looking up at the hoop during a layup or down at the ball and the goal before a penalty kick. Quantitatively, our models improve PSNR by up to 1.25\,dB on multi-view generation and from 7.75 to 12.41\,dB on exo-to-ego generation (Appendix~\ref{app:gen-benchmark}); depth accuracy is evaluated in Section~\ref{sec:depth-results}, and more examples are in Appendix~\ref{sec:qualitative}.

\subsection{Interleaved Geometric Reasoning}
\label{sec:reasoning-results}

Table~\ref{tab:reasoning-main} compares GeoUG ($\lambda=0.1$; Section~\ref{sec:adaptive-reasoning}) with open-source VLMs and DA3 on depth and distance questions, using circular accuracy.

\begin{table}[t]
  \centering
  \caption{\textbf{Depth and distance understanding.} Circular accuracy (\%); both option orders must be correct. Bold marks the best result per column. GeoUG uses adaptive interleaved reasoning and is initialized from the draft-conditioned SFT model (Section~\ref{sec:adaptive-reasoning}). DA3 answers from its predicted metric depth with the same measurement procedure as GeoUG, without a VLM. Avg.: mean of the three Overall columns.}
    \label{tab:reasoning-main}
  \begingroup
  \setlength{\tabcolsep}{3pt}
  \renewcommand{\arraystretch}{0.95}
  \fontsize{8}{9.5}\selectfont
  \begin{tabular}{@{}lcccccccc@{}}
    \toprule
    \multirow{2}{*}{Model / configuration} & \multicolumn{3}{c}{CV-Bench-3D} & \multicolumn{3}{c}{3DCueBench} & \multicolumn{1}{c}{DA-2K} & \multirow{2}{*}{Avg.} \\
    \cmidrule(lr){2-4}\cmidrule(lr){5-7}\cmidrule(lr){8-8}
    & Depth & Distance & Overall & Depth & Distance & Overall & Overall & \\
    \midrule
    BAGEL-7B-MoT~\citep{bagel} & 81.5 & 59.3 & 70.4 & 73.8 & 49.5 & 61.6 & 55.1 & 62.4 \\
    Qwen2.5-VL-7B~\citep{qwen25vl} & 80.2 & 61.5 & 70.8 & 61.8 & 35.5 & 48.6 & 45.3 & 54.9 \\
    Qwen3-VL-8B~\citep{qwen3vl} & 93.8 & 80.3 & 87.1 & 80.0 & 53.5 & 66.8 & 44.6 & 66.2 \\
    InternVL3.5-8B~\citep{internvl35} & 83.2 & 74.7 & 78.9 & 62.8 & 52.5 & 57.6 & 31.6 & 56.0 \\
    SenseNova-SI-1.4-8B~\citep{sensenova_si} & 93.5 & 85.2 & 89.3 & 81.3 & 63.0 & 72.1 & 63.9 & 75.1 \\
    SenseNova-SI-1.5-8B~\citep{sensenova_si} & 90.8 & 84.0 & 87.4 & 79.0 & 64.0 & 71.5 & 61.1 & 73.3 \\
    VST-7B-SFT~\citep{vst} & 95.5 & 90.8 & 93.2 & 83.8 & 72.0 & 77.9 & 63.5 & 78.2 \\
    \midrule
    DA3~\citep{da3} (depth readout) & 96.0 & 94.5 & 95.3 & \textbf{95.3} & \textbf{86.5} & \textbf{90.9} & 92.1 & 92.8 \\
    \midrule
    \multicolumn{1}{@{}>{\columncolor{oursblue}[0pt][\tabcolsep]}l}{GeoUG (Ours)} & \cellcolor{oursblue}\textbf{98.5} & \cellcolor{oursblue}\textbf{97.3} & \cellcolor{oursblue}\textbf{97.9} & \cellcolor{oursblue}95.0 & \cellcolor{oursblue}\textbf{86.5} & \cellcolor{oursblue}90.8 & \cellcolor{oursblue}\textbf{94.2} & \multicolumn{1}{>{\columncolor{oursblue}[\tabcolsep][0pt]}c@{}}{\textbf{94.3}} \\
    \bottomrule
  \end{tabular}
  \endgroup
\end{table}

\noindent\textbf{GeoUG outperforms the evaluated VLMs across all three benchmarks.}
Table~\ref{tab:reasoning-main} compares GeoUG with open-source VLMs and the depth estimator DA3. Using the same measurement procedure, GeoUG exceeds DA3 by 2.7 points on CV-Bench-3D and 2.1 points on DA-2K and achieves similar accuracy on 3DCueBench (90.8\% versus 90.9\%), while answering 32.6\% of 3DCueBench questions without generating depth. Supplying existing VLMs with DA3 measurements improves accuracy on CV-Bench-3D and 3DCueBench, but their performance remains below GeoUG (Appendix~\ref{sec:depth-tools}). Referring format affects models differently, with no consistent advantage for either format (Appendix~\ref{sec:visual-textual}).

\begin{figure}[p]
  \centering
  \begin{subfigure}{\linewidth}
    \centering
    \includegraphics[width=0.8\linewidth]{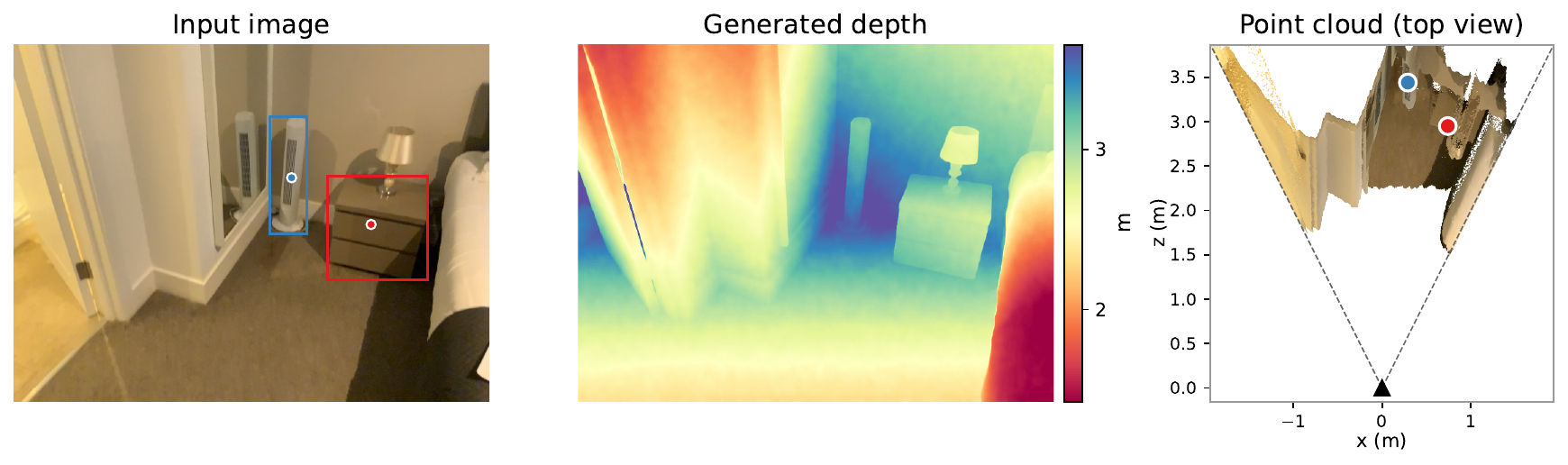}\\[2pt]
    \qualbox{
      \textbf{Question.} Which object is closer to the camera taking this photo, the \qA{nightstand} or the \qB{tower fan}? A.~nightstand \quad B.~tower fan

      \textbf{GeoUG.} To answer this, I need each object's metric distance from the camera. [boxes and points for both objects] My initial answer without depth: B.~tower fan.
      \qtok{DECIDE}\qtok{GENERATE\_DEPTH} \hfill $p_{\mathrm{gen}} = 0.93 / 0.96$

      [generated depth map, 1.48--7.28\,m] \quad \texttt{query\_point\_depth}: \qA{nightstand} 2.95\,m, \qB{tower fan} 3.44\,m.

      From the tool, \ldots\ Since 2.95 $<$ 3.44, the nightstand is closer and the tower fan is further. \textbf{Answer: A.~nightstand} \checkmark
    }
    \caption{\textbf{3DCueBench, Depth}, relative gap in $[0.10,0.16)$.}
    \label{fig:qual-d3-depth}
  \end{subfigure}\\[3pt]
  \begin{subfigure}{\linewidth}
    \centering
    \includegraphics[width=0.8\linewidth]{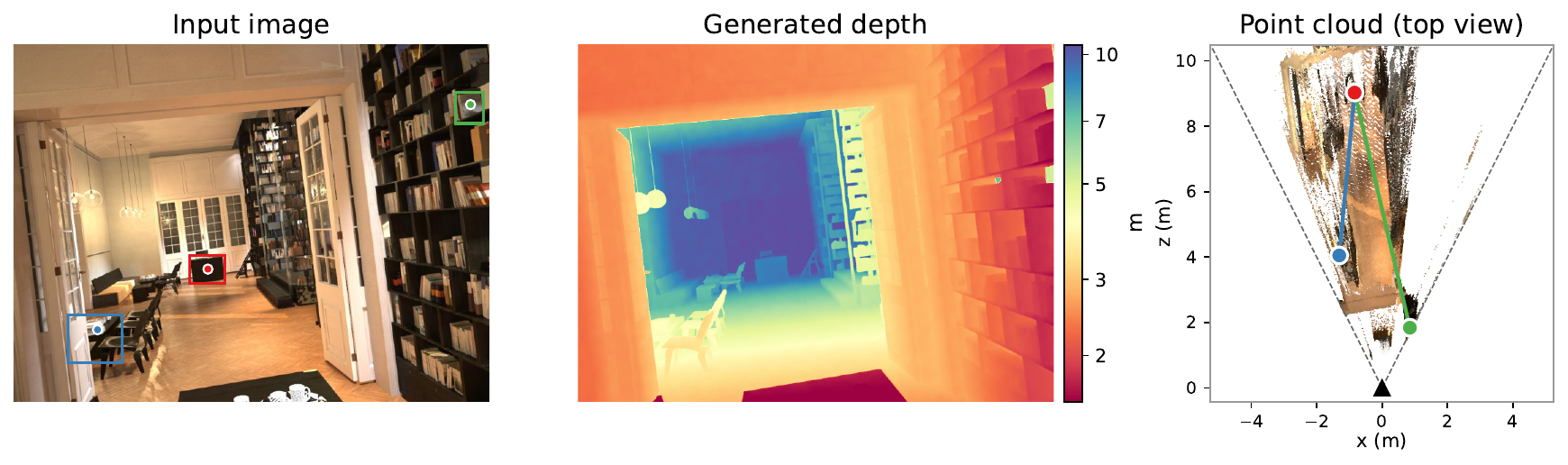}\\[2pt]
    \qualbox{
      \textbf{Question.} Estimate the real-world distances between objects in this image. Which object is closer to the \qA{desk}, the \qB{table} or the \qC{books}? A~table \quad B~books

      \textbf{GeoUG.} To answer this, I need the 3D distance from the anchor to each candidate. [a point inside each given box] My initial answer without depth: B~books.
      \qtok{DECIDE}\qtok{GENERATE\_DEPTH} \hfill $p_{\mathrm{gen}} = 0.94 / 0.95$

      [generated depth map, 0.89--12.08\,m] \quad \texttt{query\_3d\_distance}: \qA{desk}--\qB{table} 5.01\,m, \qA{desk}--\qC{books} 7.50\,m.

      This asks which object is closer to the desk \ldots\ Since 5.01 $<$ 7.50, the table is closest to the desk. \textbf{Answer: A~table} \checkmark
    }
    \caption{\textbf{CV-Bench-3D~\citep{rw_cambrian}, Distance}, objects given as boxes.}
    \label{fig:qual-cvbench}
  \end{subfigure}\\[3pt]
  \begin{subfigure}{\linewidth}
    \centering
    \includegraphics[width=0.8\linewidth]{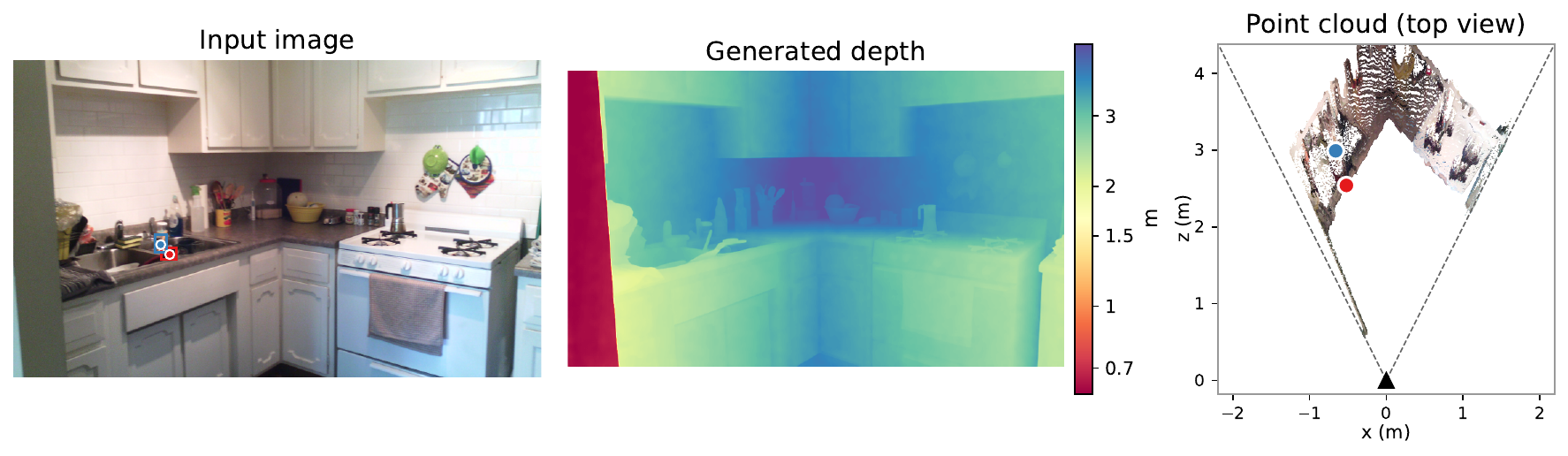}\\[2pt]
    \qualbox{
      \textbf{Question.} Which object is closer to the camera taking this photo, the \qA{first point} or the \qB{second point}? A.~first point \quad B.~second point

      \textbf{GeoUG.} To answer this, I need each object's metric distance from the camera. [a box around each point] My initial answer without depth: B.~second point.
      \qtok{DECIDE}\qtok{GENERATE\_DEPTH} \hfill $p_{\mathrm{gen}} = 0.98 / 0.93$

      [generated depth map, 0.57--4.83\,m] \quad \texttt{query\_point\_depth}: \qA{first point} 2.54\,m, \qB{second point} 2.99\,m.

      From the tool, \ldots\ Since 2.54 $<$ 2.99, the first point is closer and the second point is further. \textbf{Answer: A.~first point} \checkmark
    }
    \caption{\textbf{DA-2K~\citep{depthanythingv2}}, two nearby points on the sink.}
    \label{fig:qual-da2k}
  \end{subfigure}
  \caption{\textbf{Adaptive interleaved reasoning with GeoUG.} GeoUG drafts an answer, decides at \qtok{DECIDE} whether to generate metric depth, and measures on its own generated depth to confirm or revise the draft. In all three examples, a wrong draft is corrected under both option orders. Images show the input, the generated depth, and a top-down point cloud; $p_{\mathrm{gen}}$ is the probability of generation for both option orders. More examples are in Appendix~\ref{sec:qualitative}.}
  \label{fig:qual-reasoning}
\end{figure}

\begin{table}[t]
  \centering
  \begin{minipage}[b]{0.478\textwidth}
    \centering
    \caption{\textbf{Adaptive interleaved reasoning.} Circular accuracy (\%)
    and generation rate (Gen., \%). Pos.: \% of questions answered with the
    same letter under both option orders (mean over benchmarks). The random
    gate generates at the $\lambda=0.1$ rate for random questions.}
    \label{tab:adaptive-circular}
    \begingroup
    \setlength{\tabcolsep}{1.2pt}
    \renewcommand{\arraystretch}{0.95}
    \fontsize{7}{8.5}\selectfont
    \begin{tabular}{@{}lccccccc@{}}
      \toprule
      \multirow{2}{*}{Configuration} & \multicolumn{2}{c}{CV-Bench-3D} & \multicolumn{2}{c}{3DCueBench} & \multicolumn{2}{c}{DA-2K} & \multirow{2}{*}{Pos.} \\
      \cmidrule(lr){2-3}\cmidrule(lr){4-5}\cmidrule(lr){6-7}
      & Acc. & Gen. & Acc. & Gen. & Acc. & Gen. & \\
      \midrule
      Direct & 91.4 & 0 & 75.4 & 0 & 71.7 & 0 & 9.7 \\
      Always-generate & 97.4 & 100 & 91.3 & 100 & 94.9 & 100 & 0.1 \\
      Initial policy (SFT) & 97.2 & 16.0 & 85.4 & 25.2 & 85.1 & 28.3 & 3.5 \\
      \midrule
      Random gate & 94.4 & 50.3 & 86.1 & 67.4 & 88.5 & 72.4 & 3.2 \\
      RL, $\lambda=0.05$ & 97.5 & 58.3 & 90.6 & 74.6 & 94.7 & 78.7 & 1.0 \\
      \multicolumn{1}{@{}>{\columncolor{oursblue}[0pt][\tabcolsep]}l}{Ours ($\lambda{=}0.1$)} & \cellcolor{oursblue}97.9 & \cellcolor{oursblue}50.3 & \cellcolor{oursblue}90.8 & \cellcolor{oursblue}67.4 & \cellcolor{oursblue}94.2 & \cellcolor{oursblue}72.4 & \multicolumn{1}{>{\columncolor{oursblue}[\tabcolsep][0pt]}c@{}}{0.7} \\
      \bottomrule
    \end{tabular}
    \endgroup
  \end{minipage}\hfill
  \begin{minipage}[b]{0.506\textwidth}
    \centering
    \caption{\textbf{Ablation of the decision advantage} ($\lambda=0.1$, 20
    updates from the same initial policy and training questions). Circular
    accuracy (\%) and generation rate (Gen., \%); Avg.: mean accuracy over the
    three benchmarks. Ours: two-branch estimate with forced generation.}
    \label{tab:adaptive-ablation}
    \begingroup
    \setlength{\tabcolsep}{1.2pt}
    \renewcommand{\arraystretch}{0.95}
    \fontsize{7}{8.5}\selectfont
    \begin{tabular}{@{}lccccccc@{}}
      \toprule
      \multirow{2}{*}{Decision advantage} & \multicolumn{2}{c}{CV-Bench-3D} & \multicolumn{2}{c}{3DCueBench} & \multicolumn{2}{c}{DA-2K} & \multirow{2}{*}{Avg.} \\
      \cmidrule(lr){2-3}\cmidrule(lr){4-5}\cmidrule(lr){6-7}
      & Acc. & Gen. & Acc. & Gen. & Acc. & Gen. & \\
      \midrule
      Always-generate & 97.4 & 100 & 91.3 & 100 & 94.9 & 100 & 94.5 \\
      Initial policy (SFT) & 97.2 & 16.0 & 85.4 & 25.2 & 85.1 & 28.3 & 89.2 \\
      \midrule
      Trajectory-level & 96.6 & 44.1 & 90.8 & 66.2 & \textbf{94.1} & 71.1 & 93.8 \\
      Sampled action & \textbf{97.6} & 41.6 & 89.6 & 62.8 & 92.8 & 65.6 & 93.4 \\
      Two-branch, unforced & 97.3 & 29.2 & 89.5 & 49.3 & 92.5 & 53.0 & 93.1 \\
      \multicolumn{1}{@{}>{\columncolor{oursblue}[0pt][\tabcolsep]}l}{Ours} & \cellcolor{oursblue}\textbf{97.6} & \cellcolor{oursblue}50.7 & \cellcolor{oursblue}\textbf{91.0} & \cellcolor{oursblue}69.3 & \cellcolor{oursblue}93.9 & \cellcolor{oursblue}73.4 & \multicolumn{1}{>{\columncolor{oursblue}[\tabcolsep][0pt]}c@{}}{\textbf{94.2}} \\
      \bottomrule
    \end{tabular}
    \endgroup
  \end{minipage}
\end{table}

\noindent\textbf{Depth generation provides larger gains on harder geometric questions.}
Within the same model, always generating depth improves circular accuracy over direct answering by 6.0, 15.9, and 23.2 points on the three benchmarks (Table~\ref{tab:adaptive-circular}). On 3DCueBench, the gain reaches 23.0 points in the two hardest gap intervals and decreases to 12.5 and 5.0 points in the easier intervals (Table~\ref{tab:adaptive-difficulty}). Generation also reduces option-position bias: the fraction of questions answered with the same letter under both option orders drops from 9.7\% to 0.1\% on average.

\noindent\textbf{Generated depth corrects wrong drafts.}
Figure~\ref{fig:qual-reasoning} shows examples from the three benchmarks in which the draft answer is wrong. In each case, the policy chooses to generate depth, reads metric depths or 3D distances from it, and revises the draft to the correct answer under both option orders. More examples, including direct answers and a failure case, are in Appendix~\ref{sec:qualitative}.

\subsection{Adaptive Interleaved Reasoning}
\label{sec:adaptive-results}

\noindent\textbf{Adaptive interleaved reasoning maintains accuracy with fewer depth generations.}
With $\lambda=0.1$, the trained GeoUG generates depth for 50.3--72.4\% of questions while staying within 0.7 percentage points of always generating across all three benchmarks (Table~\ref{tab:adaptive-circular}). The initial SFT policy generates less often but achieves lower accuracy, particularly on DA-2K. Reducing $\lambda$ to 0.05 increases generation rates to 58.3--78.7\%, with accuracy changing by at most 0.5 points. On 3DCueBench, this policy stays within 1.0 point of always generating across difficulty intervals (Table~\ref{tab:adaptive-difficulty}).

\noindent\textbf{The draft-conditioned policy selects questions that benefit from generation.}
At the same generation rates, GeoUG outperforms random selection by 3.5, 4.7, and 5.7 points on CV-Bench-3D, 3DCueBench, and DA-2K, respectively (Table~\ref{tab:adaptive-circular}). This advantage holds across all difficulty intervals on 3DCueBench. After RL, fewer incorrect drafts are returned without checking, indicating that the policy learns to use depth where direct answering fails.

\noindent\textbf{Comparing generation with retaining the draft and using forced rollouts improves accuracy.}
We ablate how the draft-conditioned policy learns when to generate (Section~\ref{sec:adaptive-reasoning}). Our decision advantage in Equation~\ref{eq:gate-adv} compares generation with retaining the draft and uses forced-generation rollouts to estimate correctness after generation. Starting from the same initial policy, we evaluate after 20 RL steps at $\lambda=0.1$ (Table~\ref{tab:adaptive-ablation}). Our method achieves 94.2\% mean accuracy across the three benchmarks. Replacing $A_i^{\mathrm{gate}}$ with the trajectory-level advantage in Equation~\ref{eq:group-adv} lowers accuracy by 0.34 points; applying the draft-correctness baseline only to the sampled action lowers it by 0.81 points. Estimating $\widehat Q_{\mathrm{gen}}$ only from rollouts that chose to generate, without forced generation, lowers accuracy by 1.09 points. In this last variant, all 16 rollouts answer directly on 4\% of the training questions, leaving no estimate of $\widehat Q_{\mathrm{gen}}$ and no learning signal for those decisions. Our method generates more often at the same $\lambda$, bringing accuracy closer to always generating.

\subsection{Depth Estimation}
\label{sec:depth-results}

\noindent\textbf{GeoUG-Depth outperforms the evaluated generative depth estimators.}
We evaluate depth generation on five benchmarks using affine alignment (Section~\ref{sec:eval-protocol}, Table~\ref{tab:comp_depth_estimation}). GeoUG-Depth achieves lower AbsRel and higher $\delta_1$ than Marigold-v1.1 and Lotus on every dataset with reported results. It also outperforms Depth Anything V2 (DA2) on both metrics across all five benchmarks, while MoGe-V2 retains lower AbsRel. Qualitative examples show the generated depth maps and corresponding 3D reconstructions (Figure~\ref{fig:gen-main-depth} and Appendix~\ref{sec:appendix-qualitative}).

\begin{table}[!htp]\centering\small

\caption{\textbf{Affine-invariant depth estimation.} AbsRel $\times100$ ($\downarrow$) and $\delta_1$ in \% ($\uparrow$) after affine alignment.}\label{tab:comp_depth_estimation}

\tablestyle{2.0pt}{0.95}\fontsize{8}{9.5}\selectfont
\begin{tabular}{llcccccccccc}\toprule

\multirow{2}{*}{Type} &\multirow{2}{*}{Method} &\multicolumn{2}{c}{NYU-v2} &\multicolumn{2}{c}{KITTI} &\multicolumn{2}{c}{ETH3D} &\multicolumn{2}{c}{Sintel} &\multicolumn{2}{c}{DIODE} \\\cmidrule{3-12}

& &AbsRel &$\delta_1$ &AbsRel &$\delta_1$ &AbsRel &$\delta_1$ &AbsRel &$\delta_1$ &AbsRel &$\delta_1$ \\\midrule

\multirow{6}{*}{\rotatebox[origin=c]{90}{\textit{Dis}}} &ZoeDepth~\citep{zoedepth} &4.76 &97.30 &5.59 &95.10 &7.27 &94.20 &21.80 &69.20 &7.80 &90.90 \\

&DA V2~\citep{depthanythingv2} &4.16 &97.90 &6.77 &94.30 &4.63 &97.20 &17.10 &76.60 &5.41 &94.60 \\

&Metric3D V2~\citep{metric3dv2} &3.94 &97.60 &3.50 &98.40 &3.24 &99.00 &26.60 &71.70 &2.75 &98.70 \\

&UniDepth V2~\citep{unidepthv2} &2.96 &98.60 &3.85 &98.10 &2.95 &98.50 &13.30 &83.20 &4.05 &96.50 \\

&Depth Pro~\citep{depthpro} &3.67 &98.20 &5.12 &96.80 &4.97 &96.40 &15.80 &80.10 &4.66 &95.60 \\

&MoGe-V2~\citep{mogev2} &2.89 &98.60 &3.75 &98.10 &2.80 &99.10 &13.30 &82.50 &3.14 &97.40 \\ \midrule

\multirow{3}{*}{\rotatebox[origin=c]{90}{\textit{Gen}}} &Marigold-v1.1~\citep{ke2024marigold} &5.50 &96.40 &10.50 &90.20 &6.90 &95.70 &29.80 &78.20 &- &- \\

&Lotus~\citep{lotus} &5.40 &96.80 &8.50 &92.20 &5.90 &97.00 &- &- &9.80 &92.40 \\

&\cellcolor{oursblue}GeoUG-Depth (ours) &\cellcolor{oursblue}3.32 &\cellcolor{oursblue}98.46 &\cellcolor{oursblue}4.84 &\cellcolor{oursblue}97.27 &\cellcolor{oursblue}3.75 &\cellcolor{oursblue}97.85 &\cellcolor{oursblue}14.72 &\cellcolor{oursblue}79.65 &\cellcolor{oursblue}4.50 &\cellcolor{oursblue}96.22 \\

\bottomrule

\end{tabular}

\end{table}

\begin{table}[!htp]\centering\small
  \caption{\textbf{Depth encoding and data scaling.} Protocol and units follow Table~\ref{tab:comp_depth_estimation}; AVG is the unweighted mean across five datasets. The last row is GeoUG-Depth in Table~\ref{tab:comp_depth_estimation}.}
  \label{tab:depth_estimation_ablation}
  \tablestyle{1.7pt}{0.95}\fontsize{8}{9.5}\selectfont
  \begin{tabular}{lcccccccccccc}\toprule
    \multirow{2}{*}{Method} &\multicolumn{2}{c}{AVG} &\multicolumn{2}{c}{NYU-v2} &\multicolumn{2}{c}{KITTI} &\multicolumn{2}{c}{ETH3D} &\multicolumn{2}{c}{Sintel} &\multicolumn{2}{c}{DIODE} \\\cmidrule{2-13}
    &AbsRel &$\delta_1$ &AbsRel &$\delta_1$ &AbsRel &$\delta_1$ &AbsRel &$\delta_1$ &AbsRel &$\delta_1$ &AbsRel &$\delta_1$ \\\midrule
    single-disparity &7.47 &92.20 &3.35 &98.40 &6.61 &95.22 &4.76 &96.66 &16.68 &76.77 &5.96 &93.96 \\
    multi-disparity (MMDC) &7.24 &92.57 &3.35 &98.38 &6.51 &95.34 &4.62 &96.99 &16.31 &77.23 &5.40 &94.88 \\
    MMDC + data scaling &6.23 &93.89 &3.32 &98.46 &4.84 &97.27 &3.75 &97.85 &14.72 &79.65 &4.50 &96.22 \\
    \bottomrule
  \end{tabular}
\end{table}

\noindent\textbf{Multi-scale depth encoding improves accuracy, with further gains from more training data.}
Table~\ref{tab:depth_estimation_ablation} compares MMDC with a single-scale encoding that repeats one channel ($c=5\,\mathrm{m}$) across the three VAE channels. MMDC reduces mean AbsRel from 7.47 to 7.24, improving four datasets and matching the baseline on NYU-v2. Expanding the training data further reduces mean AbsRel to 6.23 and increases mean $\delta_1$ from 92.6\% to 93.9\%.

\section{Conclusion}
\label{sec:conclusion}

We studied how geometric generation supports spatial understanding through both learning and reasoning, using GeoUG as a unified framework.
During training, joint learning with depth and multi-view generation yields task-dependent gains.
During inference, our evaluation on 3DCueBench shows larger benefits from generated depth on questions requiring fine-grained geometric discrimination.
Building on these findings, we developed adaptive interleaved reasoning, where a draft-conditioned policy learns through reinforcement learning to balance accuracy gains against generation cost.
Across three benchmarks, this policy maintains accuracy close to or above always generating with fewer generation calls.
These results highlight two complementary roles for geometric generation: supervision for learning spatial relations and evidence acquired selectively during reasoning.

\clearpage

\bibliographystyle{plainnat}
\bibliography{main}

\clearpage

\beginappendix

\etocdepthtag.toc{mtappendix}

To support a thorough understanding of our work, a detailed Appendix has been compiled. The table of contents below serves as a guide to the specific sections of interest.

\begingroup
\etocsettagdepth{mtmain}{none}
\etocsettagdepth{mtappendix}{subsection}
\etocsettocstyle{\section*{Appendix Contents}}{}
\hypersetup{linkcolor=black}
\tableofcontents
\endgroup

\section{Qualitative Results}
\label{sec:appendix-qualitative}
\label{sec:qualitative}

\begin{figure}[!ht]
  \centering
  \includegraphics[width=\textwidth]{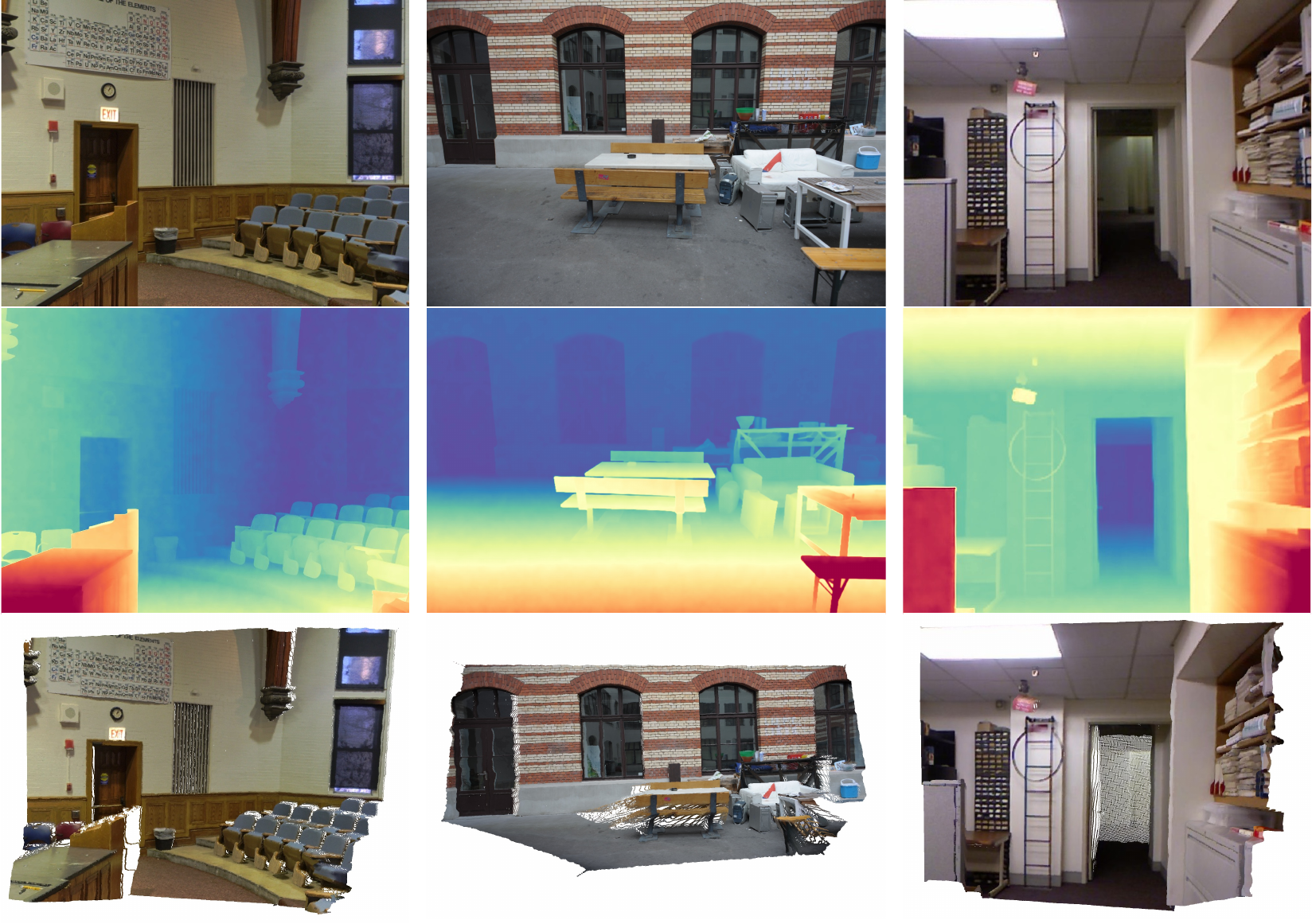}
  \caption{\textbf{Qualitative depth generation.} Rows show input images, predicted depth maps, and colored 3D reconstructions.}
  \label{fig:depth-qualitative}
\end{figure}

Figure~\ref{fig:depth-qualitative} shows predicted depth maps and their corresponding colored 3D reconstructions. The examples preserve recognizable scene layout, including the tiered seating in the lecture hall, tables in front of the courtyard facade, and the recessed doorway in the office. Depth boundaries follow many visible object contours, while the reconstructions also expose gaps and distortions near some boundaries. These examples illustrate recovered spatial structure without serving as a quantitative test of metric calibration.

\begin{figure}[t]
  \centering
  \includegraphics[width=\textwidth]{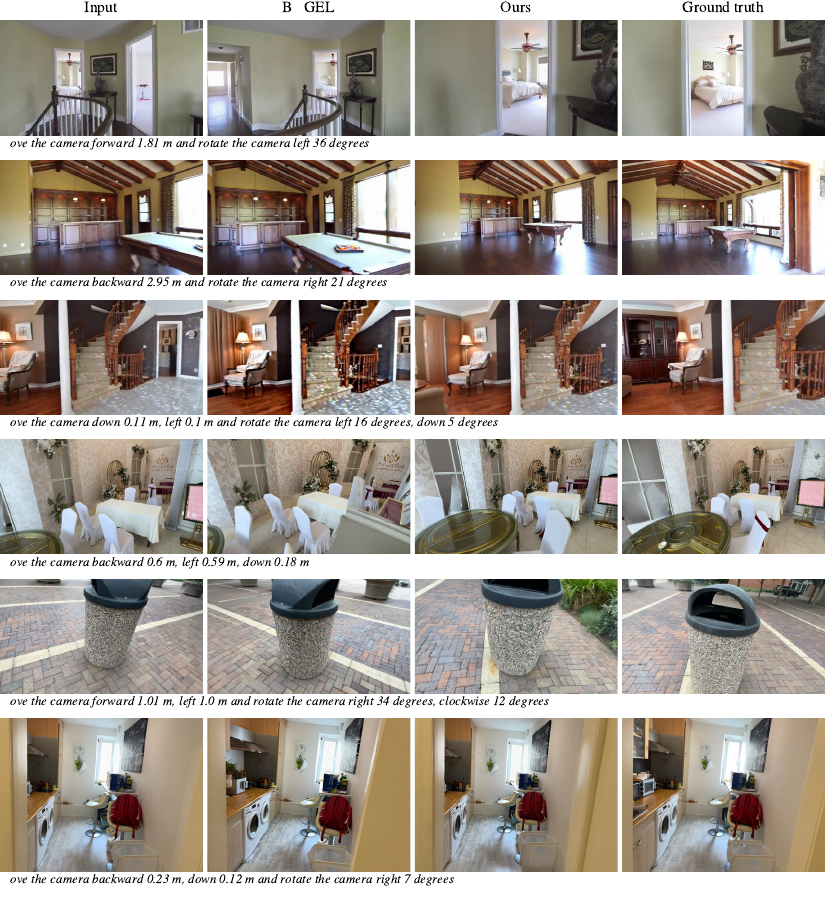}
  \caption{\textbf{Qualitative multi-view generation} with ground truth. Rows 1--3 are from RealEstate10K~\citep{re10k}, rows 4--5 from DL3DV~\citep{dl3dv}, and row 6 from ScanNet++~\citep{scannet++}. BAGEL~\citep{bagel} is the original model. The camera-motion instruction is given below each row. Ours is the multi-view co-training model (Figure~\ref{fig:cotraining}) at its final checkpoint.}
  \label{fig:qual-multiview}
\end{figure}

\begin{figure}[t]
  \centering
  \includegraphics[width=0.86\textwidth]{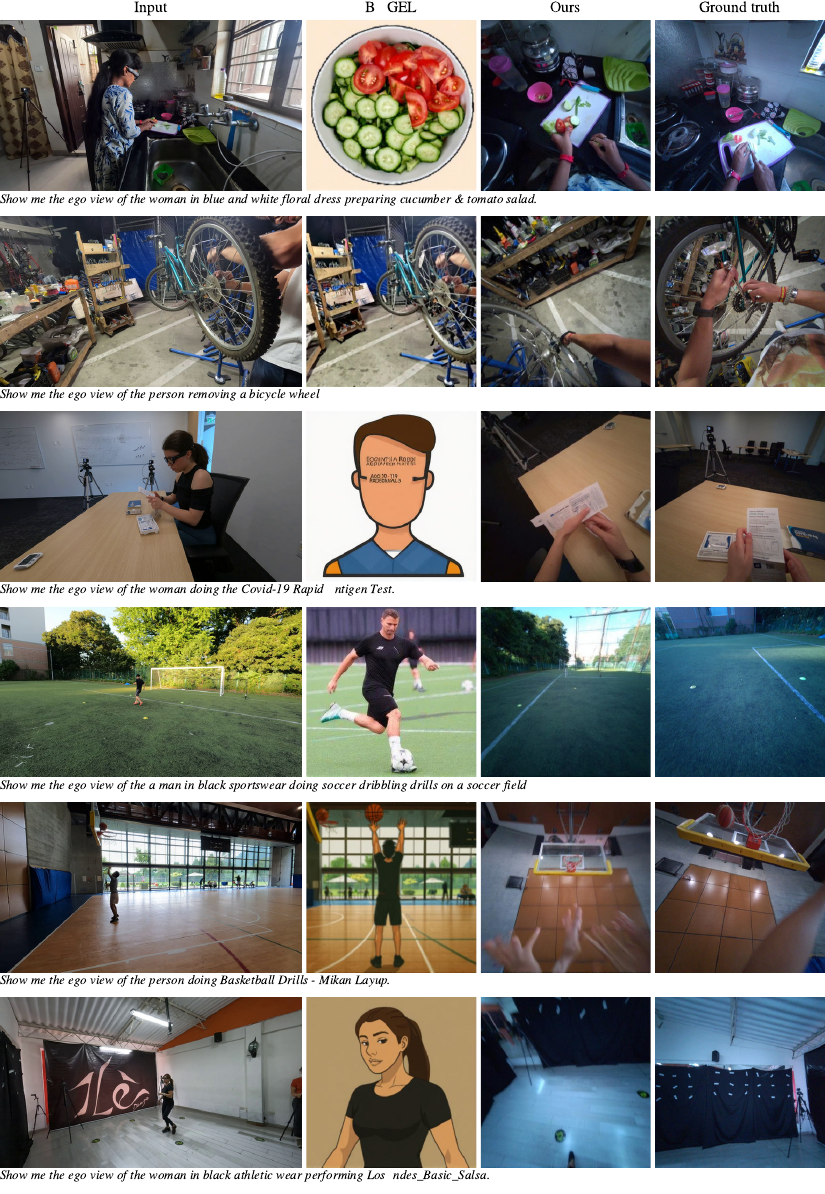}
  \caption{\textbf{Qualitative exo-to-ego generation} on Ego-Exo4D~\citep{grauman2024ego} with ground truth. Given an exo image and an instruction naming the person, the model generates that person's ego view. BAGEL~\citep{bagel} is the original model. Ours is the exo-to-ego co-training model (Figure~\ref{fig:cotraining}) at its final checkpoint. Instructions are shown verbatim.}
  \label{fig:qual-exo-ego}
\end{figure}

Figures~\ref{fig:qual-multiview} and~\ref{fig:qual-exo-ego} compare the original BAGEL~\citep{bagel} with our models on multi-view and exo-to-ego generation; quantitative results are in Tables~\ref{tab:gen-multiview} and~\ref{tab:gen-exo-ego}. The examples are chosen from random subsets of the evaluation sets (Appendix~\ref{app:gen-benchmark}): 90 multi-view pairs (30 each from RealEstate10K~\citep{re10k}, DL3DV~\citep{dl3dv}, and ScanNet++~\citep{scannet++}) and 64 of the 469 exo-to-ego pairs from Ego-Exo4D~\citep{grauman2024ego}. For multi-view generation, BAGEL often returns a view close to the input, even for instructions that move the camera by more than a metre or rotate it by more than 30\textdegree{}. Our model, trained on the pose-aligned instructions of Appendix~\ref{app:multiview-data}, moves the camera in the instructed direction and brings the expected content into view. For exo-to-ego generation, BAGEL mostly depicts the described person, or the object the person is working on, instead of rendering the scene from that person's viewpoint. Our model instead produces a first-person view, with hands, tools, and the surrounding layout roughly matching the ground truth. The generated views are not exact reconstructions: fine details, such as the arrangement of objects on the table, and the precise viewing direction often differ from the ground truth.

\begin{figure}[htbp]
  \centering
  \includegraphics[width=0.85\linewidth]{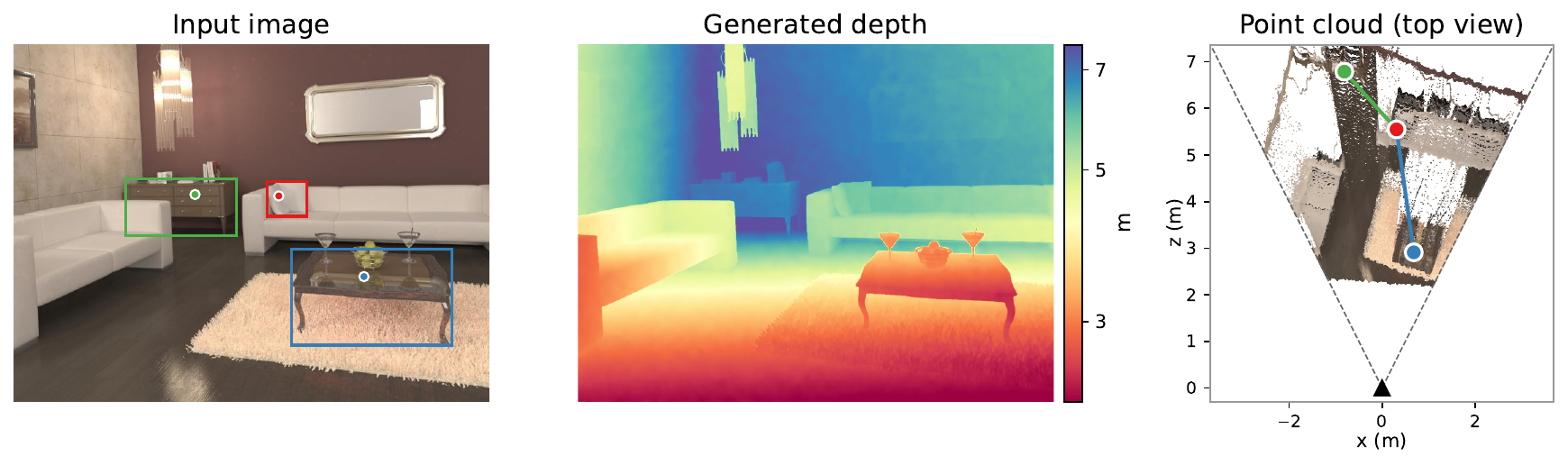}\\[3pt]
  \qualbox{
    \textbf{Question.} Estimate the real-world distances between objects in this image. Which object is closer to the \qA{white cushion}, the \qB{glass coffee table} or the \qC{wooden sideboard}? A.~glass coffee table \quad B.~wooden sideboard

    \textbf{GeoUG.} To answer this, I need the 3D distance from the anchor to each candidate. [boxes and points for the three objects] My initial answer without depth: A.~glass coffee table.
    \qtok{DECIDE}\qtok{GENERATE\_DEPTH} \hfill $p_{\mathrm{gen}} = 0.99 / 0.99$

    [generated depth map, 2.18--9.34\,m] \quad \texttt{query\_3d\_distance}: \qA{white cushion}--\qB{glass coffee table} 2.74\,m, \qA{white cushion}--\qC{wooden sideboard} 1.68\,m.

    This asks which object is closer to the white cushion --- that is the distance between objects, not the distance from the camera \ldots\ Since 1.68 $<$ 2.74, the wooden sideboard is closest to the white cushion. \textbf{Answer: B.~wooden sideboard} \checkmark
  }
  \caption{\textbf{3DCueBench, Distance}, relative gap in $[0.10,0.16)$. The draft picks the candidate nearer to the camera. After generating depth, the policy compares 3D distances between objects rather than camera depths; both option orders are answered correctly.}
  \label{fig:qual-d3-distance}
\end{figure}

\begin{figure}[htbp]
  \centering
  \begin{minipage}[t]{0.49\linewidth}
    \centering
    \includegraphics[height=3.4cm]{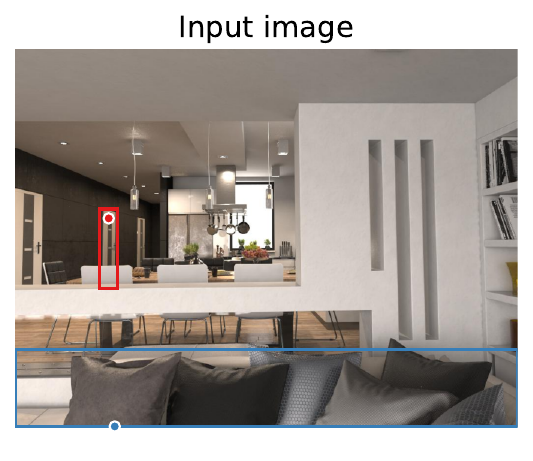}\\[3pt]
    \qualbox{
      \textbf{Question.} Which object is closer to the camera taking this photo, the \qA{door} or the \qB{pillow}? A~door \quad B~pillow

      \textbf{GeoUG.} [a point inside each given box] My initial answer without depth: B~pillow.
      \qtok{DECIDE} \hfill $p_{\mathrm{gen}} = 0.02 / 0.05$

      \textbf{Answer: B~pillow} \checkmark
    }
  \end{minipage}\hfill
  \begin{minipage}[t]{0.49\linewidth}
    \centering
    \includegraphics[height=3.4cm]{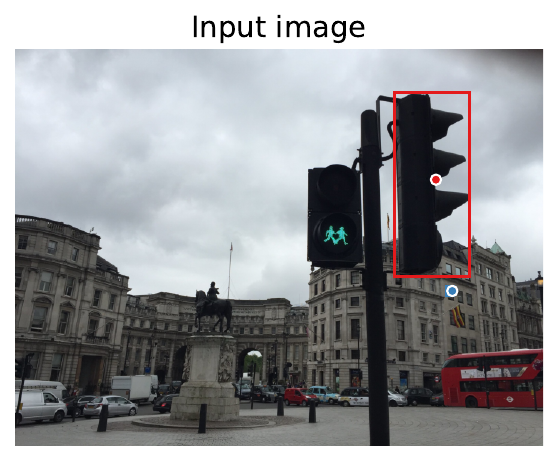}\\[3pt]
    \qualbox{
      \textbf{Question.} Which object is closer to the camera taking this photo, the \qA{first point} or the \qB{second point}? A.~first point \quad B.~second point

      \textbf{GeoUG.} [a box around each point] My initial answer without depth: A.~first point.
      \qtok{DECIDE} \hfill $p_{\mathrm{gen}} = 0.02 / 0.13$

      \textbf{Answer: A.~first point} \checkmark
    }
  \end{minipage}
  \caption{\textbf{Answering without generation.} Left: CV-Bench-3D~\citep{rw_cambrian}. Right: DA-2K~\citep{depthanythingv2}. The draft is correct and the policy answers directly after \qtok{DECIDE}; both option orders are answered correctly without generating depth.}
  \label{fig:qual-direct}
\end{figure}

\begin{figure}[htbp]
  \centering
  \includegraphics[width=0.85\linewidth]{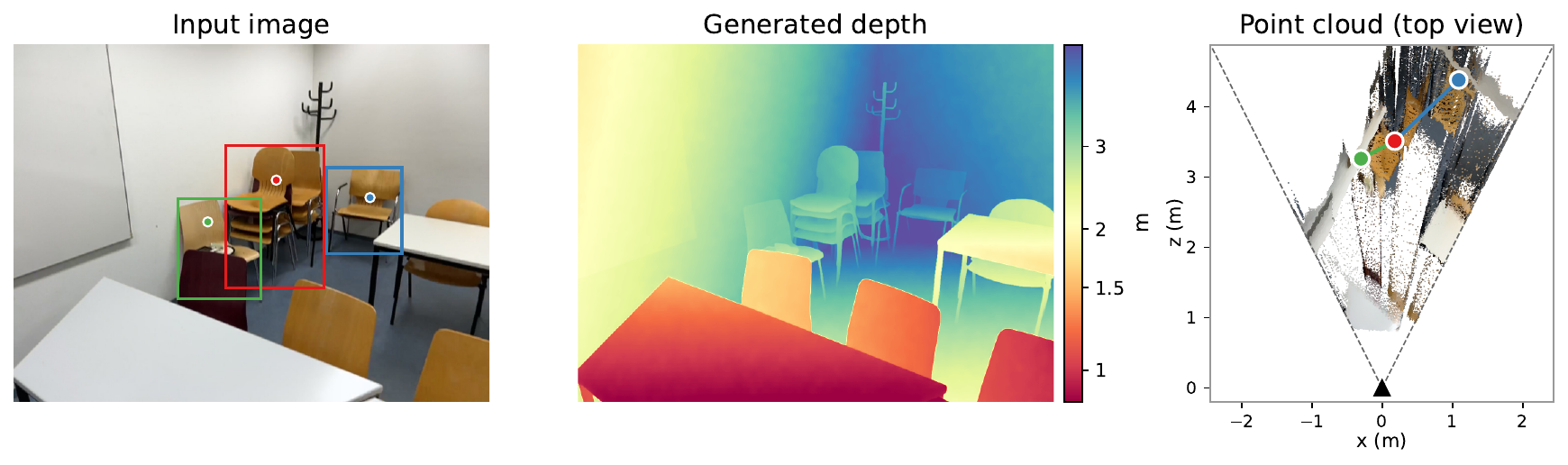}\\[3pt]
  \qualbox{
    \textbf{Question.} Estimate the real-world distances between objects in this image. Which object is closer to the \qA{stacked chairs}, the \qB{wooden armchair} or the \qC{wooden chair}? A.~wooden armchair \quad B.~wooden chair

    \textbf{GeoUG.} To answer this, I need the 3D distance from the anchor to each candidate. [boxes and points for the three objects] My initial answer without depth: A.~wooden armchair.
    \qtok{DECIDE}\qtok{GENERATE\_DEPTH} \hfill $p_{\mathrm{gen}} = 0.87 / 0.88$

    [generated depth map, 0.81--6.20\,m] \quad \texttt{query\_3d\_distance}: \qA{stacked chairs}--\qB{wooden armchair} 1.25\,m, \qA{stacked chairs}--\qC{wooden chair} 0.63\,m.

    \ldots\ Since 0.63 $<$ 1.25, the wooden chair is closest to the stacked chairs. \textbf{Answer: B.~wooden chair} $\times$ (ground truth: A)
  }
  \caption{\textbf{Failure case: 3DCueBench, Distance}, relative gap in $[0.25,0.40)$. The draft is correct, but the policy generates depth and follows the tool, whose measured distances place the wooden chair closer to the anchor. Both option orders are answered incorrectly.}
  \label{fig:qual-failure}
\end{figure}

Figure~\ref{fig:qual-reasoning} in Section~\ref{sec:reasoning-results} and Figures~\ref{fig:qual-d3-distance}--\ref{fig:qual-failure} show adaptive interleaved reasoning by GeoUG on 3DCueBench, CV-Bench-3D~\citep{rw_cambrian}, and DA-2K~\citep{depthanythingv2}. Each trace is the policy's output for the first option order, abridged: tool-call arguments are omitted and tool results are reduced to the readings the policy uses. $p_{\mathrm{gen}}$ is the probability of \qtok{GENERATE\_DEPTH} at \qtok{DECIDE}, reported for both option orders. The depth maps are those the policy generated. The point clouds back-project them with the same pinhole camera that the tool uses when no intrinsics are given (focal length equal to the longer image side, centered principal point), and are shown from above with the camera at the bottom; points more than 0.3\,m above the highest queried object are removed so that the ceiling does not hide the scene. Colored markers are the object centroids returned by the tool; for distance questions, the lines connect the anchor to each candidate.

\FloatBarrier

\section{Details and Analysis of MMDC}
\label{app:mmdc}

\subsection{Depth sensitivity}
\label{app:hparam}
\begin{figure}[htbp]
  \centering
  \includegraphics[width=0.80\linewidth]{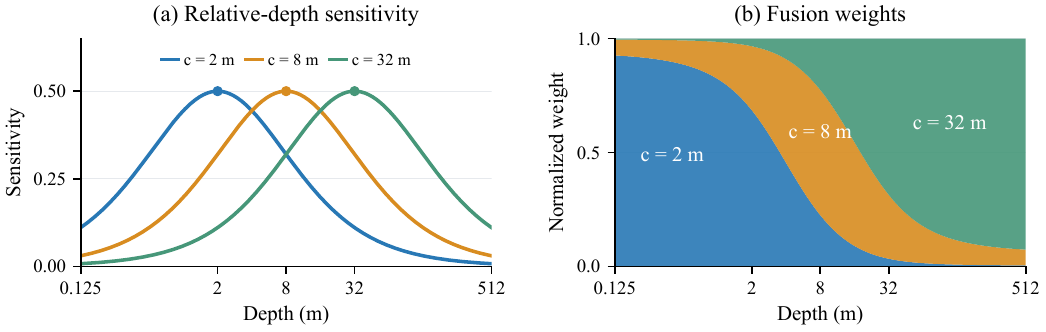}
  \caption{\textbf{MMDC sensitivity and fusion.} (a) Channel sensitivity. (b) Normalized fusion weights. Both are evaluated on noiseless encoded depths.}
  \label{fig:mmdc-sensitivity}
\end{figure}

For a channel with pivot $c_i$, let $u_i=\log(d/c_i)$ measure proportional
changes in depth. The encoding's sensitivity and local error amplification are
\begin{equation}
  s_i(d)=\frac{\partial Y_i}{\partial u_i}
  =\frac{2c_i d}{(d+c_i)^2}=\frac{1-Y_i^2}{2},\qquad
  \frac{\Delta d}{d}\approx\frac{\Delta Y_i}{s_i(d)}.
  \label{eq:mmdc-sensitivity}
\end{equation}
A first-order expansion of the inverse gives the second expression:
the same encoding error produces a larger relative depth error when
sensitivity is low. Sensitivity peaks at $d=c_i$
and decreases towards both smaller and larger depths
(Figure~\ref{fig:mmdc-sensitivity}a).

The encoding can also be written as $Y_i=\tanh(u_i/2)$.
Changing the pivot therefore translates the sensitivity profile along the
log-depth axis. Our geometrically spaced pivots $(2,8,32)\,\mathrm{m}$
place these profiles at equal log-depth intervals, providing overlapping
precision ranges. The pivots remain fixed across datasets.

\subsection{Fusion weights}
\label{app:weight}
We use the sensitivity analysis to derive a local approximation to the
fusion weights. At one pixel, assume small, independent, zero-mean encoding
errors with equal variance $\sigma^2$: $\hat Y_i=Y_i+\varepsilon_i$.
Away from the clipping boundaries, \eqref{eq:mmdc-sensitivity} gives
\begin{equation}
  v_i\equiv\operatorname{Var}\!\left(\frac{\hat d_i-d}{d}\right)
  \approx\frac{\sigma^2}{s_i(d)^2}
  =\frac{4\sigma^2}{(1-Y_i^2)^2}.
  \label{eq:app-var-d}
\end{equation}
For a weighted estimate $\hat d=\sum_i a_i\hat d_i$, with
$\sum_i a_i=1$, the relative-error variance is approximately
$\sum_i a_i^2v_i$. Minimizing it yields inverse-variance weights:
\begin{equation}
  a_i=\frac{v_i^{-1}}{\sum_j v_j^{-1}},\qquad
  w_i\propto s_i(d)^2\propto(1-Y_i^2)^2.
  \label{eq:app-weight}
\end{equation}
Channels with higher sensitivity receive greater weight
(Figure~\ref{fig:mmdc-sensitivity}b).
In practice, we substitute the clipped predictions $\hat Y_i$ for the
unknown noiseless values $Y_i$, giving \eqref{eq:decode}.
This motivates the decoder as a first-order approximation under the stated
noise assumptions, rather than an exact optimum for arbitrary generation errors.

\subsection{Implementation details}
Depths and pivots are expressed in metres. The reference implementation
marks positive depths as valid and intersects them with an optional validity
mask; invalid entries are filled with zero, with the mask retained separately.
Before inversion, predicted channels are clipped to
$[-1+10^{-4},1-10^{-4}]$. Fusion weights are floored at $10^{-12}$, and
the decoded depth has a configurable upper bound of $10^4\,\mathrm{m}$ by
default. These operations keep the inverse and weighted average numerically
well-defined near saturation.

\FloatBarrier

\section{Details of Adaptive Interleaved Reasoning}
\label{sec:app-adaptive}

\paragraph{Pilot evaluation protocols.}
Both pilot studies report ordinary (non-circular) accuracy.
Figure~\ref{fig:difficulty-by-task} compares separately trained Direct and
Always-generate checkpoints. Within each task, both datasets use five fixed
relative-gap bins, drawn at equal spacing. Depth bins start at
0.10, 0.16, 0.25, 0.40, and 0.55; distance bins at
0.10, 0.25, 0.40, 0.60, and 0.80. Intervals are left-closed and
right-open, except that the last includes 1.00. Bins with fewer than ten
questions are shaded and their estimates omitted; bands are pointwise
95\% Wilson intervals.

In Figure~\ref{fig:rl-pilot}b, we rank the 800 3DCueBench questions by
generation probability (descending), or by the ground-truth relative gap
(ascending), and use generated answers for the top fraction and direct
answers for the remainder. We report the fraction of the always-generate
gain retained,
\[
  \frac{\mathrm{Acc}-\mathrm{Acc}_{\mathrm{direct}}}
       {\mathrm{Acc}_{\mathrm{always}}-\mathrm{Acc}_{\mathrm{direct}}}.
\]
Direct and generated answers come from the same checkpoint within each
curve; the before-answer and draft-conditioned policies use different
checkpoints and supervised training. Ties are averaged under uniform
random ordering. Random selection follows the diagonal in expectation.
Figure~\ref{fig:rl-pilot}c plots generation rates during RL with reward
$c-\lambda a$. Each update uses 16 questions; the plotted rates use an
exponential moving average over questions with smoothing factor 0.97.
The proposed policy uses the 1500-step supervised initialization, and its
forced-generation rollouts are excluded from the plotted rate. The before-answer
and draft-conditioned settings differ in model, training questions, and credit
assignment; this comparison does not isolate decision position alone.

\paragraph{Pilot analysis of the before-answer policy.}
The before-answer policy is trained on a format in which the decision
token follows a short, largely templated reasoning preamble and both branches are
supervised with equal frequency; the before-answer results in
Fig.~\ref{fig:rl-pilot}b and c use this policy.
Its generation probability took two values across the 800 3DCueBench questions
(0.469 and 0.500), and the resulting gate was 0.6 points above a random gate at
the same generation rate. Hidden states at the decision token were nearly
identical across questions (median pairwise cosine 0.998), and a linear probe
predicting whether depth helps reached an AUC of 0.536, compared with 0.513 for
shuffled labels. With symmetric supervision, the maximum-likelihood gate is a
constant. Under reinforcement learning (Fig.~\ref{fig:rl-pilot}c), $\lambda=0$
drove the final generation rate to 0.96, and $\lambda\in\{0.05,0.07,0.1\}$ to at
most 0.005; evaluated accuracy followed the random-gate line (residual standard
deviation 0.025, below the per-point standard error of 0.05). In these runs,
changing the cost mainly changes the overall generation rate, without
producing an effective question-selection policy.
Ranking by the ground-truth gap in Fig.~\ref{fig:rl-pilot}b uses the draft-conditioned
policy's outcomes; each curve is normalized by its own policy's direct and
always-generate accuracies (before-answer 82.5\% and 90.0\%, draft-conditioned
81.0\% and 91.1\%).

\paragraph{Advantage scale.}
Standard GSPO~\citep{gspo} divides by the within-group standard deviation $s_r$
(Eq.~\eqref{eq:group-adv}). With a generation cost, this normalization removes
the magnitude of $\lambda$ in a second way. In a group where every rollout answers
correctly but only some generate, $r_i=1-\lambda a_i$ and
\[
  A_i^{\mathrm{group}}=\frac{-\lambda(a_i-\bar a)}{\lambda s_a+\epsilon}
  \approx-\frac{a_i-\bar a}{s_a},
\]
where $s_a$ is the standard deviation of $a_i$ within the group. Any $\lambda>0$
then yields the same full-magnitude advantage against generating, whereas in groups
whose correctness varies the cost only shifts the advantages by $\lambda/s_r$.
Because all-correct groups are frequent, this term dominates the decision's update.
We therefore divide all advantages by a constant $s=0.5$, the standard deviation of
a Bernoulli variable with mean 0.5. Every advantage is then linear in $\lambda$,
and groups whose rewards differ only by the cost receive proportionally small updates.

\paragraph{Failure of trajectory-level advantages.}
With one group-normalized advantage per trajectory, $A_i=(r_i-\bar r)/s$,
consider a group in which every rollout generates. Then $r_i=c_i-\lambda$ for all
$i$, so $A_i=(c_i-\bar c)/s$ does not depend on $\lambda$: the cost no longer
pushes the decision, whose update reflects only which generating rollouts
happened to answer correctly. Once the policy generates on nearly every rollout,
nothing pushes it to stop. In addition, a rollout that generates although its draft was
already correct receives a positive advantage whenever $\bar r<1-\lambda$, and
wrong draft tokens share the positive advantage when depth corrects them.
In our first runs with this objective (500-step initialization), the generation
rate rose from 0.64 to 0.99 within about ten updates for $\lambda=0.05$, and
drafts drifted (18 of 63 drafts without an answer letter after eight updates,
versus 7 of 64 at the start).

\paragraph{Segment-wise advantages.}
Each trajectory is split into four segments: grounding tokens (localization),
draft tokens (from ``My initial answer without depth:'' through \texttt{<DECIDE>}),
the decision token, and execution tokens (depth request, geometric queries, final
answer). With $G$ rollouts of one question, let $\mathcal{D}$ be the rollouts that
generated and $\mathcal{F}\subset\mathcal{D}$ those forced to generate.
\begin{itemize}
  \item Grounding tokens: $c_i-\bar c_{\mathcal{D}}$ for $i\in\mathcal{D}$ and $0$
  otherwise.
  \item Draft tokens: $0$, so that reinforcement learning does not sharpen the draft
  confidence the decision reads; in the supervised stage, the draft answer also
  receives no loss.
  \item Decision token: \eqref{eq:gate-adv} for $i\notin\mathcal{F}$ and $0$ for
  $i\in\mathcal{F}$. $\widehat Q_{\mathrm{gen},i}$ is the mean correctness of forced
  rollouts with the same draft correctness as rollout $i$, falling back to all
  forced rollouts if that cell is empty. For exact conditional branch values,
  $Q_{\mathrm{dir},i}=c_{d,i}$ and
  $V_i=p_i(Q_{\mathrm{gen},i}-\lambda)+(1-p_i)Q_{\mathrm{dir},i}$ yield the two
  action advantages in \eqref{eq:gate-adv}. Their expected unclipped
  score-function update is
  $p_i(1-p_i)(Q_{\mathrm{gen},i}-\lambda-c_{d,i})$ with respect to the decision
  logit. In practice, the pooled forced rollouts approximate the conditional
  generation value. They are selected uniformly at random to avoid relying
  only on rollouts whose policy chose generation; grouping by draft
  correctness does not guarantee an unbiased estimate for every distinct
  reasoning prefix. $p_i$ is read from the
  sampled token's log-probability; the two action tokens carry at least 99.6\% of
  the probability mass at the decision token.
  \item Execution tokens: $c_i-\bar c_{\mathcal{D}}$ for $i\in\mathcal{D}$; for direct
  rollouts the final answer equals the draft and the advantage is $0$.
\end{itemize}
All advantages are divided by a constant scale of 0.5. The policy loss uses
clipped sequence-level importance ratios~\citep{gspo}, one per segment, with each
segment weighted by its token count.

\begin{figure}[t]
  \centering\footnotesize
  \begin{minipage}[c]{0.27\linewidth}
    \includegraphics[width=\linewidth]{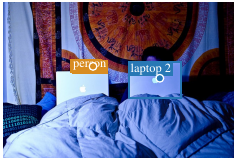}
  \end{minipage}\hfill
  \begin{minipage}[c]{0.70\linewidth}
    \textbf{(a)} \textit{Which object is closer to the camera, the laptop 2 or the person?}\\
    To answer this, I need each object's metric distance from the camera. [boxes and points]
    My initial answer without depth: A. laptop 2 \texttt{<DECIDE></think>} A. laptop 2
  \end{minipage}

  \vspace{6pt}
  \includegraphics[width=\linewidth]{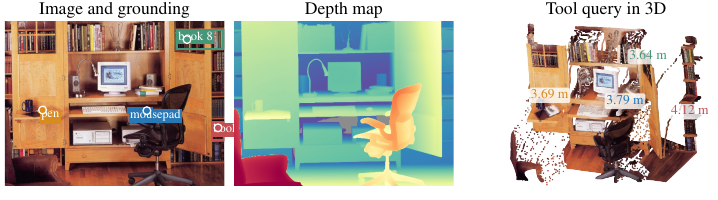}\\[-1pt]
  \begin{minipage}{\linewidth}
    \textbf{(b)} \textit{Which one is farthest from the camera: the mousepad, the pen, the book 8, or the book 12?}\\
    To answer this, I need each object's metric distance from the camera. [boxes and points]
    My initial answer without depth: C. book 8 \texttt{<DECIDE><GENERATE\_DEPTH>} [depth map]
    [query the depths of the four objects: 3.79\,m, 3.69\,m, 3.64\,m, 4.12\,m]
    The book 12 is farthest from the camera. \texttt{</think>} D. book 12
  \end{minipage}

  \vspace{6pt}
  \includegraphics[width=\linewidth]{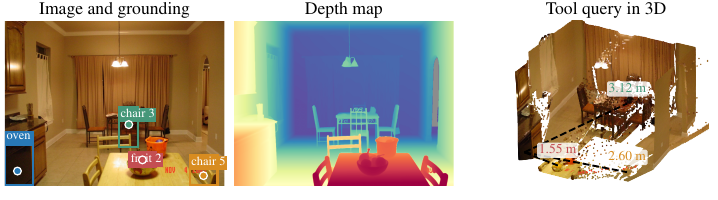}\\[-1pt]
  \begin{minipage}{\linewidth}
    \textbf{(c)} \textit{Which of these is closest to the oven: the fruit 2, the chair 3, or the chair 5?}\\
    To answer this, I need the 3D distance from the anchor to each candidate. [boxes and points]
    My initial answer without depth: C. chair 5 \texttt{<DECIDE><GENERATE\_DEPTH>} [depth map]
    [query distances from the oven: chair 5, 2.60\,m;
    chair 3, 3.12\,m; fruit 2, 1.55\,m]
    The fruit 2 is closest to the oven. \texttt{</think>} A. fruit 2
  \end{minipage}
  \caption{\textbf{Training sequences of the draft-then-decide format.}
  (a) A confident, correct draft answered directly (form (i)).
  (b, c) Wrong drafts checked with depth (form (iii)) for a camera-depth and an
  object-to-object question. Panels show the image with the queried boxes and points,
  the depth map the model learns to generate, and the query in 3D: the depth map is
  back-projected with the camera intrinsics, and markers show the positions returned by
  the geometric query, labeled with their depths (b) or distances to the anchor (c).
  The draft answer after ``My initial answer without depth:'' receives no loss;
  geometric queries and their results are summarized in brackets.}
  \label{fig:rl-data}
\end{figure}

\paragraph{Draft-conditioned supervised data.}
Supervised sequences are built from comparison questions generated by our data
engine on COCO~\citep{coco} and ADE20K~\citep{ade20k} images with DA3~\citep{da3} depth. Drafts come from an intermediate
checkpoint of the same before-answer training run, which also initializes the
draft-conditioned supervised stage. For each question, we feed the image, the question,
and the reasoning prefix of the training sequence followed by ``My initial answer
without depth:'', and take the option letter with the highest next-token
probability. Drafts thus reflect the beliefs of the model that supervised training
starts from, and greedy selection matches greedy decoding at inference. Drafts are
wrong on about 22\% of the questions. Each question
yields up to three sequences sharing the prefix up to \texttt{<DECIDE>}
(Fig.~\ref{fig:rl-data}): (i) a correct draft answered directly; (ii) a correct
draft followed by depth generation, geometric queries, and the answer; and (iii) a
wrong draft followed by depth generation and the correct answer. Draft tokens receive
no loss. The decision token is supervised only in (i), as stopping, and in (iii), as
generating; in (ii) it is masked, so that sequences with identical context do not
pull it toward 0.5, while the rest of (ii) teaches how to use depth. A wrong draft is
never answered directly, and generation follows both correct and wrong drafts, so
that generating does not simply mean changing the answer. Keeping one quarter of the
direct sequences balances the supervised decisions. After removing near ties
(relative gap below 5\%), the set contains 262K questions and 312K sequences.

\paragraph{Reinforcement-learning training questions.}
Reinforcement learning uses questions generated by our data engine on ScanNet++~\citep{scannet++}
(indoor) and nuScenes~\citep{nuscenes} (outdoor) images. Labels come from sensor depth measured on
object surfaces rather than box centers. We remove questions that name annotation
artifacts or large structural surfaces (walls, floors, ceilings, doors, windows),
whose distance is ill-defined. For object-to-object questions, we keep only those
where 2D cues, such as image-plane distances between objects, point to a wrong
answer; object-to-camera questions are kept regardless. We keep at most two
questions per image, preferring those with the most misleading 2D cues within each
question type, and sample a pool balanced between indoor and outdoor scenes and
across question types, split by image into training and validation. The 800
training questions contain one question per image and are balanced across the two
sources, four question types, nearer versus farther targets, and answer letters.

\paragraph{Training setup.}
The draft-conditioned supervised stage trains for 3000 steps. The policy is initialized
from its 1500-step checkpoint, which has the most accurate drafts among 500--3000 steps. We train
LoRA~\citep{lora} adapters (rank 32) on the language model with learning rate $5\times10^{-5}$,
16 rollouts per question of which 4 are forced to generate, and 16 questions per
update, for one epoch over the 800 questions (50 updates).

\FloatBarrier

\section{Data}
\label{app:data}
Table~\ref{tab:data source} summarizes the sources of instruction data used for different tasks. Below, we outline the data processing details for specific tasks.

\begin{table}[!htp]
\begin{tabular}{l| p{9cm}}\toprule
Task &Data sources \\\midrule
Relative depth estimation &Hypersim~\citep{hypersim}, VirtualKitti~\citep{virtualkitti}, UrbanSyn~\citep{gomez2025all}, UnrealStereo4K~\citep{Tosi2021CVPR}, Structured3D~\citep{zheng2020structured3d}, Spring~\citep{mehl2023spring}, MidAir~\citep{fonder2019mid}, MatrixCity~\citep{li2023matrixcity}, IRS~\citep{wang2021irs}, GTA-SfM~\citep{wang2020flow}, EDEN~\citep{le2021eden}, DynamicReplica~\citep{karaev2023dynamicstereo}, Bedlam~\citep{black2023bedlam}, MVS-Synth~\citep{huang2018deepmvs} \\\midrule
Metric depth estimation &Scannet++~\citep{scannet++}, Matterport3D~\citep{chang2017matterport3d}, Arkitscenes~\citep{baruch2021arkitscenes}, Taskonomy~\citep{zamir2018taskonomy}, Wildrgbd~\citep{xia2024rgbd}, Cityscapes~\citep{cordts2016cityscapes}, Waymo~\citep{waymo}, AgiBotWorld~\citep{agibot} \\\midrule
Multi-view generation &RealEstate10K~\citep{re10k}, DL3DV-10K~\citep{dl3dv} \\\midrule
Exo-to-ego generation &Ego-Exo4D~\citep{grauman2024ego} \\\midrule
Visual instruction following &LLaVA-OneVision~\citep{llavaov}, VST~\citep{vst} \\
\bottomrule
\end{tabular}
\caption{Data source statistics.}
\label{tab:data source}
\end{table}

\subsection{Metric Depth Data}

\noindent \textbf{Real Data Refinement.}
Real training data for the metric depth estimation are commonly derived from either LiDAR scans or Structure-from-Motion (SfM) reconstructions: LiDAR suffers from sensor–camera synchronization errors that corrupt depth along object boundaries, while SfM fails on reflective surfaces, thin structures, and sharp discontinuities, leaving holes or oversmoothed regions. A depth estimator trained on synthetic data avoids both issues—color and depth are exactly registered by construction and fine local geometry is preserved—but it predicts relative depth only up to an unknown affine transform. We therefore use such a relative depth estimator~\citep{da3} as a teacher and treat its output as a dense pseudo label that repairs the real measurement, after first removing this affine ambiguity.

Concretely, we fit a single scale $a$ and shift $b$ that take the relative depth $\tilde{D}$ into the metric frame of the real depth $D$. Let $m_i\in\{0,1\}$ be a validity mask indicating which pixels carry a usable real measurement. The two scalars are determined by a masked least-squares fit,
\begin{equation}
    (\hat{a},\hat{b})
    \;=\;\arg\min_{a,\,b}\;\sum_{i=1}^{N} m_i\,\bigl(a\,\tilde{D}_i + b - D_i\bigr)^{2},
    \qquad
    \hat{D}\;\triangleq\;\hat{a}\,\tilde{D}+\hat{b},
    \label{eq:ss_align}
\end{equation}
which has a unique closed-form solution as long as the valid pixels span more than one depth value. The aligned depth $\hat{D}$ matches the metric scale of $D$ on the supervised pixels while inheriting the dense, sharp geometry of the relative depth estimator, and it is $\hat{D}$ that enters our subsequent training objectives.

\begin{figure}[!t]
  \centering
  \includegraphics[width=1.0\linewidth]{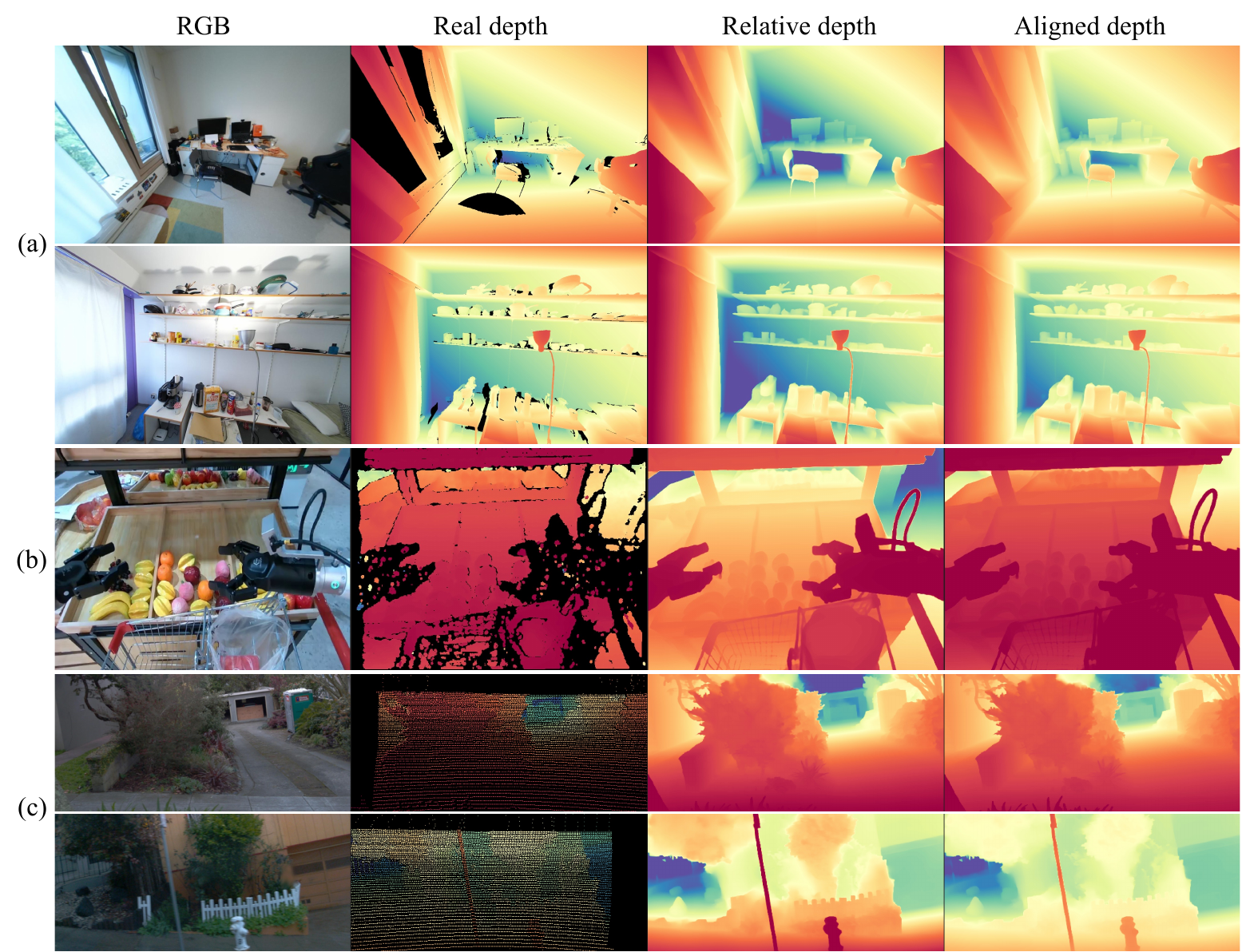} 
    \caption{%
        \textbf{Relative-to-metric depth alignment.} (a) ScanNet++~\citep{scannet++}; (b) AgiBotWorld~\citep{agibot}; (c) Waymo~\citep{waymo}. 
    }
  \label{fig:vis_metric_depth_align}
\end{figure}

\subsection{Multi-view Data}
\label{app:multiview-data}

\begin{figure}[!t]
  \centering
  \includegraphics[width=1.0\linewidth]{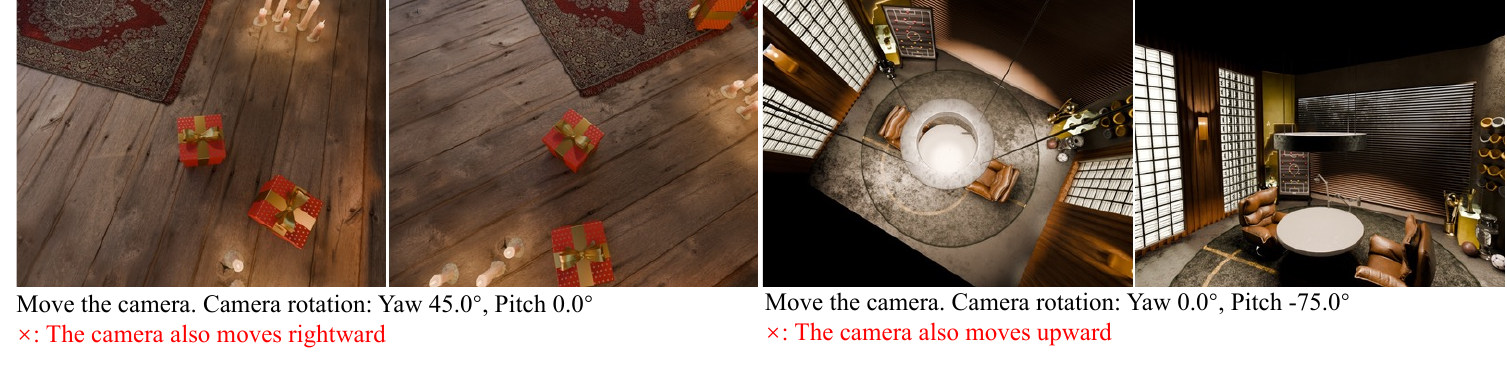} 
    \caption{%
        \textbf{SpatialEdit instruction examples}~\citep{spatialedit}. Translation is omitted from the camera instructions.
    }
  \label{fig:mv_generation_spatial_edit}
\end{figure}

\begin{figure}[htbp]
  \centering
  \includegraphics[width=0.65\linewidth]{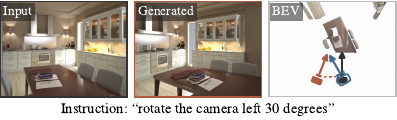}
  \caption{\textbf{Instruction misalignment.} Asked to rotate left by 30\textdegree{}, BAGEL instead rotates the camera about 21\textdegree{} right and moves it about 1\,m left. BEV: input (black), instructed (blue), and generated (red) cameras.}
  \label{fig:multiview-misalignment}
\end{figure}

\begin{figure}[!t]
  \centering
  \includegraphics[width=1.0\linewidth]{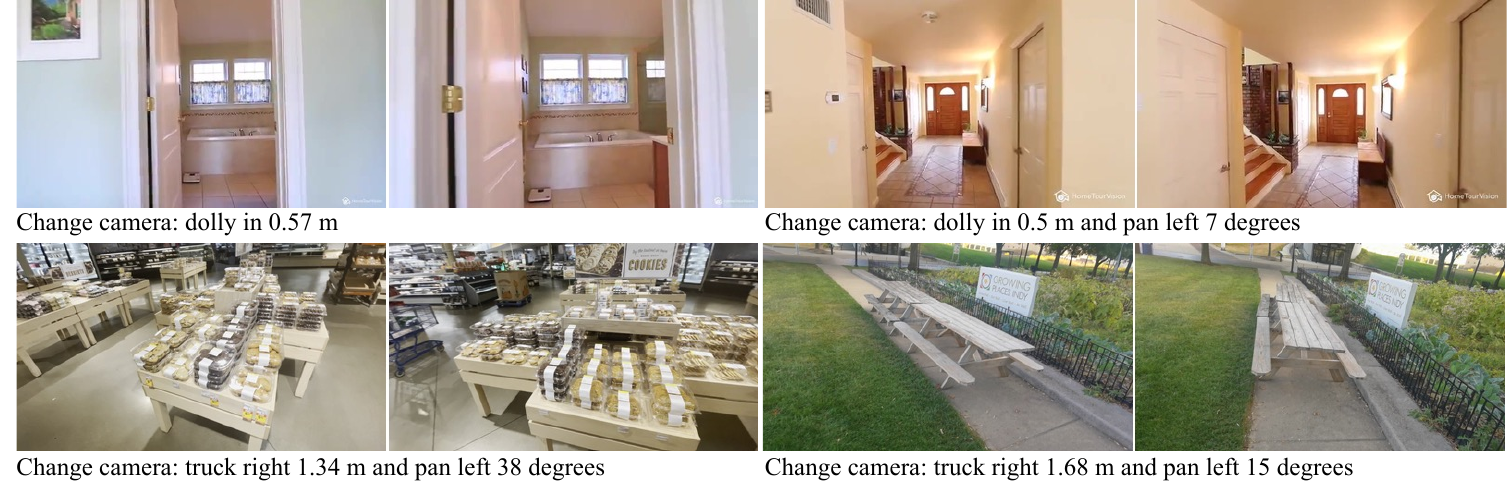} 
    \caption{%
        \textbf{Our multi-view samples.} Instructions describe both rotation and translation after scale unification.
    }
  \label{fig:mv_generation_vis}
\end{figure}

\noindent\textbf{Instruction--View Alignment.}
Camera-motion instructions should describe the transformation between the
source and target views. An instruction that specifies only rotation can
omit translation present in the paired images. For example, the SpatialEdit
pair in Figure~\ref{fig:mv_generation_spatial_edit} is labeled as
\emph{Pan Left}, although the camera also undergoes \emph{Truck Right}.
The original BAGEL shows the same inconsistency at generation time:
Figure~\ref{fig:multiview-misalignment} shows an output that rotates right and
shifts sideways when a left rotation is requested. We hypothesize that such
misaligned pairs in existing training data contribute to this failure, since
they associate the stated motion with visual changes caused by a different
transformation. We therefore construct instructions from camera poses so that
they account for both rotation and translation.
Figure~\ref{fig:mv_generation_vis} shows the resulting training pairs.

\noindent \textbf{Scale Unification.}
Since the camera poses in RealEstate10K~\citep{re10k} and DL3DV~\citep{dl3dv} are extracted using SfM-like methods, they inherently lack a consistent physical metric scale. This scale ambiguity prevents us from directly mapping camera translations to text instructions. For example, ``move forward by 1 meter" could equal 1 unit in one scene but 10 units in another scene. To address this inconsistency, we re-estimate all camera poses using DA3~\citep{da3} to unify the scale information. Thus, for a video, we get the world-to-camera poses: $\{P_i = (R_i, t_i)\}_{i=1}^{N}$.

\noindent \textbf{Instruction Construction.}
Our pose-based data engine expresses relative motion in the source camera's
coordinate frame and uses its rotation and translation to construct the
instruction. Each training example consists of a source image, the resulting
motion instruction, and the paired target image.
Given a video paired with the camera poses, we label
every short segment of a clip with a few words from a simple vocabulary
\emph{Dolly / Truck / Pedestal / Pan / Tilt / Roll} (each signed), plus
\emph{Stay} for near-static cases. The motion between frames $a$ and
$b$, written in camera-$a$'s local frame, is
\begin{equation}
\label{eq:local-delta}
\mathbf{d}_{ab} = R_{a}(c_{b}-c_{a}),\quad
\boldsymbol{\omega}_{ab} = \mathrm{Euler}_{yxz}(R_{b} R_{a}^{\top}),\quad
c_{k} = -R_{k}^{\top} t_{k};
\end{equation}
each component is thresholded to decide whether a given atomic motion
fires (e.g.\ a sufficiently positive $d_z$ triggers \emph{Dolly In}).

A single frame gap is brittle: short gaps are dominated by pose noise,
long ones blur neighbouring motions. We therefore smooth the trajectory
with a quaternion-aware EMA (flipping the sign to stay on the shorter
arc) and repeat the atomic extraction over a grid
$(\Delta,\alpha)\in\mathcal{G}$ of gaps and smoothing strengths, with
thresholds rescaled by $\sqrt{\Delta}$ so that every scale contributes
comparable votes. For each unit interval we pool these votes, drop the
loser of any opposite pair (e.g.\ \emph{Pan\,Left}/\emph{Right}), and
keep the top label alone if it leads by $3\times$, otherwise every label
within half of the leader, capped at four.

Neighbouring intervals with identical labels are merged into segments
$[a,b)$; to obtain multi-scale supervision we further enumerate every
sub-interval inside each segment. For each one we recompute the exact
magnitudes $\mathbf{m}$ from the \emph{unsmoothed} poses and project
them onto the voted labels (e.g.\ the magnitude of \emph{Dolly\,In} is
simply $d_z$). Three light filters then clean the result: a
\emph{consistency} check removes labels whose recomputed direction
contradicts the vote, an \emph{absolute} check drops per-frame
magnitudes below the sensor-noise level, and a \emph{relative} check,
applied separately to translation and rotation, discards secondary
motions weaker than $\rho$ times the dominant one. Segments whose
strongest motion still falls below a global cut-off are dropped
entirely, leaving the final supervision
$\{([a,b),\mathcal{Y},\mathbf{m})\}$ summarized in Algorithm~\ref{alg:p2i}.

\begin{algorithm}[t]
\caption{Pose $\rightarrow$ Instruction}
\label{alg:p2i}
\small
\DontPrintSemicolon
\KwIn{Poses $\{P_i\}$; grid $\mathcal{G}$; thresholds $\tau,\tau^{\text{abs}},\rho,\tau^{\min}$}
\KwOut{Labeled segments $\mathcal{S}=\{([a,b),\mathcal{Y},\mathbf{m})\}$}
$P^{\star} \leftarrow \mathrm{QuatEMA}(P;\alpha_0)$ \tcp*{reference trajectory}
\lForEach{$(\Delta,\alpha)\in\mathcal{G}$}{%
  $L^{\Delta,\alpha} \leftarrow \mathrm{Thresh}\!\left(\Delta\mathrm{Pose}(\mathrm{QuatEMA}(P;\alpha),\Delta);\,\tau\sqrt{\Delta}\right)$%
}
$\{y_i\} \leftarrow \mathrm{TopK}\circ\mathrm{Mutex}\circ\mathrm{Vote}\!\left(\bigcup_{\Delta,\alpha} L^{\Delta,\alpha}\right)$\;
$\{[a_k,b_k),\mathcal{Y}_k\} \leftarrow \mathrm{Merge}(\{y_i\})$\;
\ForEach{sub-interval $[a,b)\subseteq[a_k,b_k)$}{
  $\mathbf{m} \leftarrow \mathrm{Project}\!\left(\Delta\mathrm{Pose}(P^{\star}_a,P^{\star}_b),\,\mathcal{Y}_k\right)$\;
  $\mathcal{Y} \leftarrow \mathrm{Filter}_{\mathrm{cons}\wedge\mathrm{abs}\wedge\mathrm{rel}}\!\left(\mathcal{Y}_k,\mathbf{m};\tau^{\text{abs}},\rho\right)$\;
  \lIf{$\mathcal{Y}\neq\emptyset \ \wedge\ \max|\mathbf{m}|\geq\tau^{\min}$}{%
    $\mathcal{S}\leftarrow\mathcal{S}\cup\{([a,b),\mathcal{Y},\mathbf{m})\}$%
  }
}
\Return $\mathcal{S}$
\end{algorithm}

We sub-sample with stride $s_0\!=\!3$ and set
$\mathcal{G}\!=\!\{1,3,5\}\!\times\!\{0.03,0.05,0.1\}$,
$(\tau_t,\tau_r)\!=\!(0.01,1.5^{\circ})$,
$(\tau_t^{\text{abs}},\tau_r^{\text{abs}})\!=\!(0.006,0.35^{\circ})$,
$\rho\!=\!0.3$, and $(\tau_t^{\min},\tau_r^{\min})\!=\!(0.01,1.5^{\circ})$;
rotation thresholds further carry a scaling constant $\kappa\!=\!1.8$.

\begin{figure}[!t]
    \centering
    \begin{subfigure}{\textwidth}
        \centering
        \includegraphics[width=1.0\linewidth]{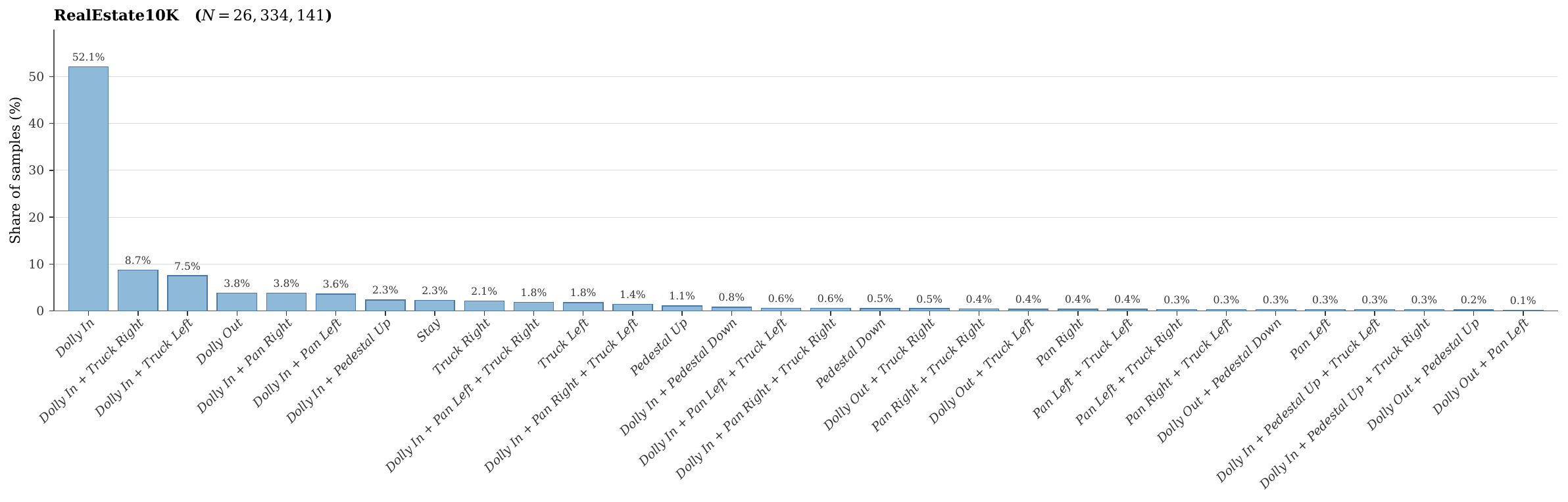} 
        \caption{Camera instruction distribution of RealEstate10K~\citep{re10k}}
        \label{fig:inst_distribution_plot_top_combos_re10k}
    \end{subfigure}
    \vspace{1em} 
    \begin{subfigure}{\textwidth}
        \centering
        \includegraphics[width=1.0\linewidth]{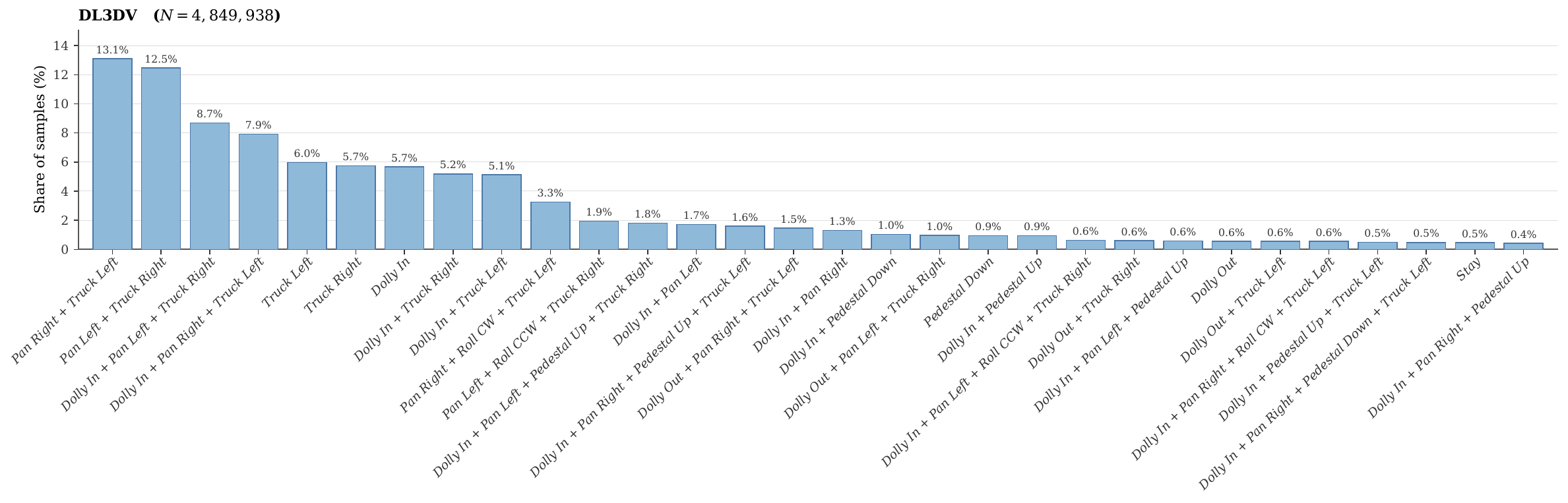}
        \caption{Camera instruction distribution of DL3DV~\citep{dl3dv}}
        \label{fig:inst_distribution_plot_top_combos_dl3dv}
    \end{subfigure}
    \caption{Camera instruction distribution in RealEstate10K~\citep{re10k} and DL3DV~\citep{dl3dv} dataset.}
    \label{fig:mv_inst_distribution}
\end{figure}

\noindent \textbf{Instruction Distribution.}
We converted the data from RealEstate10K~\citep{re10k} and DL3DV~\citep{dl3dv} into camera instructions and observed noticeable biases in their distributions, as shown in Figure~\ref{fig:mv_inst_distribution}. For instance, the dominant instruction in RealEstate10K is \emph{Dolly\,In}, since this dataset largely consists of room-tour videos. In contrast, DL3DV is dominated by \emph{Pan} and \emph{Truck}. This is because DL3DV is designed for 3D reconstruction, which typically involves moving the camera around a central object.

To address this issue, we sample the data to maintain a balanced distribution across different instructions. Additionally, we noticed that combined camera motions are much more frequent than single camera motions. Therefore, we resampled the single-motion data to train our model using a curriculum learning strategy. Specifically, we first train the model to learn single camera motions before introducing combined motions.

\subsection{Exo-to-Ego Generation Data}
\label{app:exo-ego-data}

\begin{figure}[!t]
  \centering
  \includegraphics[width=1.0\linewidth]{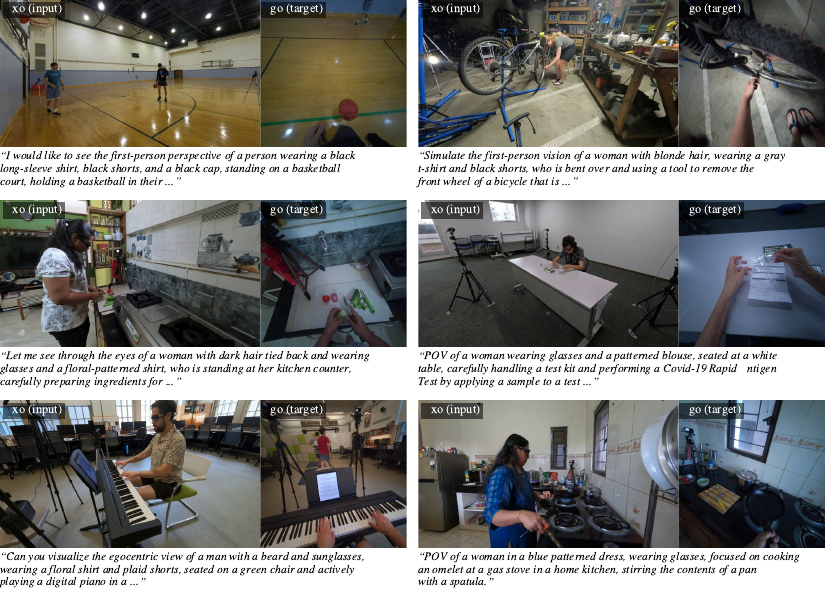}
    \caption{%
        \textbf{Exo-to-ego training pairs.} Each pair shows the exo input and the synchronized ego target from Ego-Exo4D~\citep{grauman2024ego}; the instruction below each pair is truncated.
    }
  \label{fig:exo_ego_data}
\end{figure}

Exo-to-ego generation is a form of multi-view generation that requires perspective taking: the target viewpoint is specified by another observer in the scene rather than by a camera-motion instruction, and the model must imagine what that observer sees. We build exo-to-ego pairs from Ego-Exo4D~\citep{grauman2024ego}, which records each take with static third-person (exo) cameras and a head-mounted first-person (ego) camera. Each pair consists of an exo frame and the ego frame captured at the same time step. A VLM describes the camera wearer in the exo frame, including appearance, action, and surroundings, and turns the description into an instruction that asks for that person's first-person view. We remove pairs whose instruction contains unparsed VLM output, leaked templates, or references to markers that are not drawn on the image. The resulting set contains 171,558 pairs covering sports, bike repair, cooking, health care, music, and dance. The ego target is resized to $512\times512$. Figure~\ref{fig:exo_ego_data} shows examples.

The task requires the model to locate the described person in the exo view, infer where that person is looking, and render the scene from the new viewpoint. The ego target therefore contains the objects in front of the person, often at a large viewpoint change from the exo camera, rather than the person themselves.

\FloatBarrier

\section{Benchmark}

\subsection{Multi-view and Exo-to-Ego Generation}
\label{app:gen-benchmark}

\noindent\textbf{Multi-view generation.} The benchmark contains 1,500 pairs, 500 each from RealEstate10K~\citep{re10k}, DL3DV~\citep{dl3dv}, and ScanNet++~\citep{scannet++}. Each pair consists of a source view, a target view, and a camera-motion instruction built from the relative pose of the two views with the procedure of Appendix~\ref{app:multiview-data}, for example ``Move the camera forward 1.81 m and rotate the camera left 36 degrees''. The model generates the target view from the source view and the instruction.

\noindent\textbf{Exo-to-ego generation.} The benchmark contains 469 pairs from Ego-Exo4D~\citep{grauman2024ego}. Each pair consists of an exo frame, the synchronized ego frame, and an instruction of the form ``Show me the ego view of $\langle$person description$\rangle$''. The model generates the $512\times512$ ego view.

\noindent\textbf{Metrics.} We compare each generated image with the ground-truth target using PSNR, SSIM, and LPIPS~\citep{zhang2018unreasonable} with an AlexNet backbone. Generated images are resized to the target resolution before scoring. Generation uses 50 denoising steps with the default classifier-free guidance of BAGEL~\citep{bagel} (text scale 4.0, image scale 2.0).

Tables~\ref{tab:gen-multiview} and~\ref{tab:gen-exo-ego} compare the original BAGEL with our models, which are the same models shown in Figures~\ref{fig:qual-multiview} and~\ref{fig:qual-exo-ego}. On multi-view generation, our model improves all three metrics on every source, with the largest gain on RealEstate10K (+1.25\,dB PSNR, $-0.071$ LPIPS). On exo-to-ego generation, BAGEL scores much lower (7.75\,dB PSNR), consistent with its tendency to depict the described person rather than that person's view (Figure~\ref{fig:qual-exo-ego}). Our model raises PSNR to 12.41\,dB and lowers LPIPS from 0.839 to 0.670. These pixel and perceptual metrics reward matching the exact target view, and absolute values remain low because a single source image and instruction do not fully determine the target.

\begin{table}[!ht]
\centering
\small
\caption{\textbf{Multi-view generation benchmark.} PSNR (dB), SSIM, and LPIPS between generated and ground-truth target views; the number of pairs is given in parentheses.}
\label{tab:gen-multiview}
\setlength{\tabcolsep}{3pt}
\resizebox{\linewidth}{!}{%
\begin{tabular}{lcccccccccccc}
\toprule
& \multicolumn{3}{c}{RealEstate10K (500)} & \multicolumn{3}{c}{DL3DV (500)} & \multicolumn{3}{c}{ScanNet++ (500)} & \multicolumn{3}{c}{All (1,500)} \\
\cmidrule(lr){2-4}\cmidrule(lr){5-7}\cmidrule(lr){8-10}\cmidrule(lr){11-13}
Model & PSNR$\uparrow$ & SSIM$\uparrow$ & LPIPS$\downarrow$ & PSNR$\uparrow$ & SSIM$\uparrow$ & LPIPS$\downarrow$ & PSNR$\uparrow$ & SSIM$\uparrow$ & LPIPS$\downarrow$ & PSNR$\uparrow$ & SSIM$\uparrow$ & LPIPS$\downarrow$ \\
\midrule
BAGEL & 13.68 & 0.411 & 0.463 & 12.05 & 0.317 & 0.580 & 13.75 & 0.700 & 0.518 & 13.16 & 0.476 & 0.520 \\
Ours & \textbf{14.93} & \textbf{0.452} & \textbf{0.392} & \textbf{12.41} & \textbf{0.321} & \textbf{0.543} & \textbf{14.22} & \textbf{0.712} & \textbf{0.497} & \textbf{13.85} & \textbf{0.495} & \textbf{0.477} \\
\bottomrule
\end{tabular}}
\end{table}

\begin{table}[!ht]
\centering
\small
\caption{\textbf{Exo-to-ego generation benchmark.} PSNR (dB), SSIM, and LPIPS between generated and ground-truth ego views on 469 Ego-Exo4D pairs.}
\label{tab:gen-exo-ego}
\begin{tabular}{lccc}
\toprule
Model & PSNR$\uparrow$ & SSIM$\uparrow$ & LPIPS$\downarrow$ \\
\midrule
BAGEL & 7.75 & 0.312 & 0.839 \\
Ours & \textbf{12.41} & \textbf{0.398} & \textbf{0.670} \\
\bottomrule
\end{tabular}
\end{table}


\subsection{3DCueBench}
\label{app:d3}

Section~\ref{sec:eval-d3} introduces the tasks and cue-balancing design.
Here we provide construction details, audit procedures, and supplementary results.

\subsubsection{Data and question format}

\begin{figure}[t]
\centering
\newcommand{\slot}[1]{{\normalfont\itshape\color{black!55}$\langle$#1$\rangle$}}
\fcolorbox{black!25}{black!3}{\parbox{0.975\linewidth}{\ttfamily\footnotesize\frenchspacing
Consider the relationship in the 3D world. Which is closer to \slot{target}?\\
(A) \slot{object A} (represented by the point <point>\slot{x} \slot{y}</point>)\\
(B) \slot{object B} (represented by the point <point>\slot{x} \slot{y}</point>)\\
Please select the correct answer from the options above.\\[3pt]
{\normalfont\footnotesize Depth: \slot{target} = the camera.\\
Distance: \slot{target} = \slot{anchor} (represented by the point \texttt{<point>}\slot{x}~\slot{y}\texttt{</point>}).}}}

\vspace{6pt}
\includegraphics[width=\linewidth]{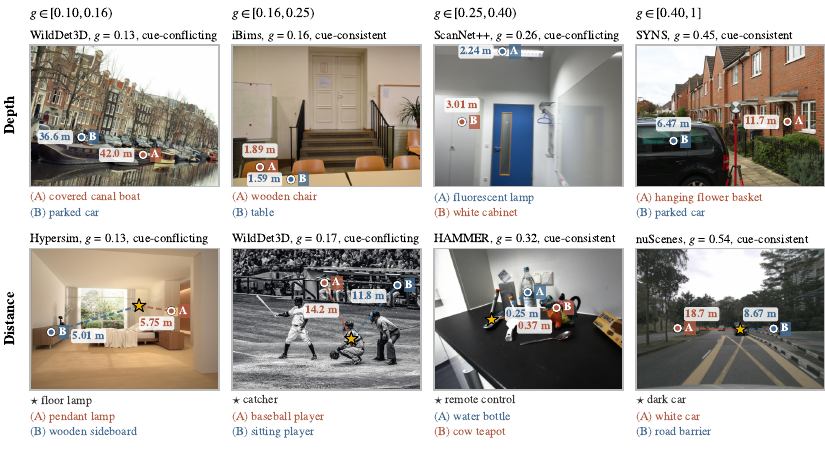}
\caption{\textbf{3DCueBench examples.} Top: the question template shared by all questions; $(x,y)$ are point coordinates normalized so that the longer image side spans 0--1000, and the image is supplied without marks. Bottom: one question per gap interval for each task. Under each image, $\star$ gives \slot{anchor} and (A)/(B) give \slot{object A}/\slot{object B}; the correct option is shown in blue. Numbers in the images are ground-truth camera depths (Depth) or 3D distances to the anchor (Distance).}
\label{fig:d3-examples}
\end{figure}

The benchmark contains 400 Depth and 400 Distance questions from ten datasets
(Table~\ref{tab:d3-sources}). The two tasks are sampled, balanced, and audited
independently. Each question has two answer options; objects are identified by
a short model-generated description and a point in the question text.
Coordinates are normalized so that the image's longer side spans $0$--$1000$.
Distance questions identify the anchor in the same format. Images supplied
to models contain no added target markers; the markers in the figures are
for illustration only. Figure~\ref{fig:d3-examples} shows the question template and one question
per gap interval for each task.

\begin{table}[h]
\centering\small
\caption{\textbf{3DCueBench sources.} Domains follow the indoor/outdoor balancing assignment.}
\label{tab:d3-sources}
\begin{tabular}{@{}l>{\raggedright\arraybackslash}p{0.84\linewidth}@{}}
\toprule
Domain & Datasets \\
\midrule
Indoor  & CA-1M~\citep{ca1m}, ScanNet++~\citep{scannet++}, Hypersim~\citep{hypersim}, iBims~\citep{ibims}, Booster~\citep{booster}, HAMMER~\citep{hammer} \\
Outdoor & nuScenes~\citep{nuscenes}, WildDet3D~\citep{wilddet3d} (in-the-wild, spans both), Waymo~\citep{waymo}, SYNS~\citep{syns} \\
\bottomrule
\end{tabular}
\end{table}

\subsubsection{Sampling and cue selection}

\paragraph{Gap intervals.}
Using the relative gap $g$ defined in Equation~\ref{eq:relative-gap}, we sample
four intervals: $[0.10,0.16)$, $[0.16,0.25)$, $[0.25,0.40)$, and $[0.40,1]$.
Each interval contains 100 Depth and 100 Distance questions, with equal
indoor and outdoor representation within each task.

\paragraph{Ground-truth sensitivity.}
The lower cutoff avoids comparisons at which different depth definitions
frequently disagree. We compare the 3D box-center depth with the 10th
percentile and median of depth within the instance mask. In the smallest-gap
CA-1M group, all three definitions agree on 44.7\% of pairs, while the two
surface-based definitions agree on 75.7\%. These discrepancies indicate
sensitivity to the depth convention; they do not by themselves establish
which convention gives the correct label.

\paragraph{Cue signatures and pairing.}
We retain candidates only when the selected task-specific 2D rules agree
with one another. Each retained question is therefore \emph{cue-consistent}
(all selected rules favor the answer) or \emph{cue-conflicting} (all favor
the other option). We pair the two types within the same source and gap
interval, preserving the balance within each source and domain.
Depth cues use vertical image-plane relationships, including box-bottom
height; Distance cues use image-plane proximity to the anchor.
The point-coordinate relation used in the qualitative analysis below is
reported separately from the box-based cue signature.

\subsubsection{CV-Bench-3D audit}

The rule accuracies in Figure~\ref{fig:3dcuebench-examples}(c) are computed
from the bounding boxes and answer labels of 600 Depth and 600 Distance
questions of CV-Bench-3D~\citep{rw_cambrian}. For Depth, the rule selects the object with the larger box-bottom
$y$ coordinate; for Distance, it selects the candidate whose box center has
the smaller Euclidean image-plane distance to the anchor's box center.
These are deterministic rule predictions, not VLM outputs.

For the difficulty audit, we match the question boxes to Omni3D test
annotations~\citep{omni3d} from Hypersim~\citep{hypersim}, SUN RGB-D~\citep{sunrgbd}, and nuScenes~\citep{nuscenes} by minimizing the sum of
absolute differences between box coordinates within each source image.
Camera-space box centers provide the depths and anchor distances used to
compute $g$. None of the 1,200 questions has $g<0.16$; median gaps are 0.49
for Depth and 0.76 for Distance, and 75\% and 98.8\%, respectively, fall in
$g\geq0.40$.

\subsubsection{Blind-text audit}
\label{app:d3-blind}

To test for residual cues in the question text, we evaluate Qwen2.5-VL-7B~\citep{qwen25vl}
without image inputs and Qwen2.5-7B-Instruct~\citep{qwen25} on both option orders.
Their accuracies range from 48.9\% to 50.4\%. On Distance, their predictions
agree with the coordinate-proximity rule on 99.6\% and 100.0\% of examples,
respectively. Gradient-boosted trees over numeric question fields achieve
56.1\% on Depth and 53.5\% on Distance under scene-level cross-validation.
We keep scenes disjoint across the probe's training and evaluation folds:
splitting by image instead inflates accuracy by 5.5 percentage points.
These probe results measure residual text-only predictability rather than
an upper bound on all possible shortcuts.

\subsubsection{Results and failure analysis}
\label{app:d3-results}

We report supplementary results for Seed-2.0-Lite~\citep{seed2} and GPT-5.5~\citep{gpt55} in
Table~\ref{tab:d3-results}. These use ordinary accuracy, rather than the
question-level circular accuracy in Table~\ref{tab:reasoning-main}.
Depth accuracy decreases as the relative gap shrinks, reaching 75.0\% and
62.6\% in the hardest interval. Distance accuracy is lower overall and
varies less consistently with the gap.

GPT-5.5 uses the \texttt{gpt-5.5-2026-04-24} version. Excluding its 12 unparseable answers can change the evaluated interval sizes from the balanced construction.

\begin{table}[t]
\centering\small
\caption{\textbf{3DCueBench results} of Seed-2.0-Lite~\citep{seed2} and GPT-5.5~\citep{gpt55}. Ordinary accuracy (\%) by gap interval and cue group. All: all questions of the task; agr.: agreement with the selected 2D rule. Twelve unparseable GPT-5.5 answers are excluded.}
\label{tab:d3-results}
\setlength{\tabcolsep}{2.2pt}
\begin{tabular}{llcccccccc}
\toprule
& & \multicolumn{4}{c}{accuracy by interval of $g$} & & \multicolumn{2}{c}{cue group} & rule \\
\cmidrule(lr){3-6}\cmidrule(lr){8-9}\cmidrule(lr){10-10}
Model & Task & $[0.10,0.16)$ & $[0.16,0.25)$ & $[0.25,0.40)$ & $[0.40,1]$ & all & Consistent & Conflicting & agr. \\
\midrule
Seed-2.0-Lite & Depth    & 75.0 & 88.0 & 94.0 & 95.0 & 88.0 & 92.5 & 83.5 & 54.5 \\
Seed-2.0-Lite & Distance & 56.0 & 54.0 & 69.0 & 79.0 & \textbf{64.5} & 92.0 & \textbf{37.0} & \textbf{77.5} \\
GPT-5.5 & Depth    & 62.6 & 70.4 & 87.9 & 90.9 & 78.0 & 86.2 & 69.8 & 58.2 \\
GPT-5.5 & Distance & 62.0 & 59.2 & 66.3 & 71.1 & \textbf{64.6} & 93.4 & \textbf{35.7} & \textbf{78.8} \\
\bottomrule
\end{tabular}
\end{table}

\begin{figure}[t]
\centering
\includegraphics[width=0.7\linewidth]{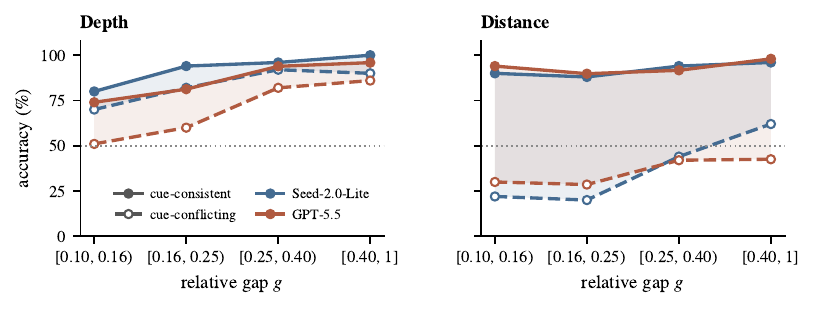}
\caption{\textbf{Sensitivity to 2D cues.} Ordinary (non-circular) accuracy of Seed-2.0-Lite and GPT-5.5 on cue-consistent (solid) and cue-conflicting (dashed) questions in each gap interval. Shading spans the difference for each model; the dotted line marks chance.}
\label{fig:d3-shortcut}
\end{figure}

\paragraph{Performance by cue group.}
On Distance, both models reach at least 92\% on cue-consistent questions but score
37.0\% and 35.7\% on cue-conflicting questions
(Figure~\ref{fig:d3-shortcut}), falling to 22\% and 30\% in the hardest
interval. Their predictions agree with the selected rule on 77.5\% and
78.8\% of questions. This asymmetry is consistent with substantial reliance
on the 2D cues, which aggregate accuracies of 64.5\% and 64.6\% obscure.
Depth shows lower model-to-rule agreement, at 54.5\% and 58.2\%.

\begin{figure}[t]
\centering
\includegraphics[width=\linewidth]{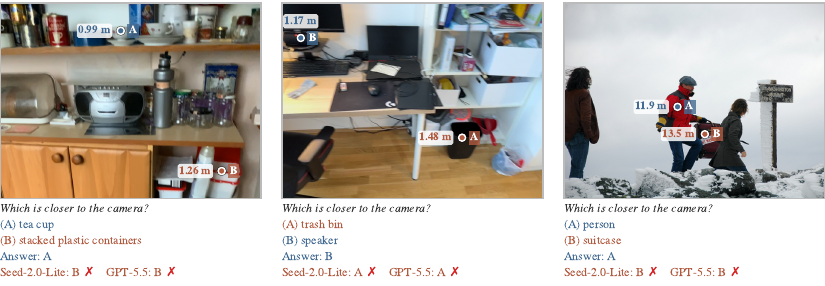}
\caption{\textbf{Counterexamples to vertical-position cues.} In each Depth question the correct option (blue) is the higher point in the image yet nearer to the camera, and both models choose the other, farther option. Under each image, a red cross after a model's answer marks it as incorrect. Numbers in the images are ground-truth camera depths; marks and model answers are shown for the reader only.}
\label{fig:d3-cases}
\end{figure}

\paragraph{What the failures look like.}
Figure~\ref{fig:d3-cases} shows Depth questions in which the higher target
point is nearer to the camera. Grouping questions by point height, rather
than the box-based signature in Table~\ref{tab:d3-results}, gives accuracies
of 82.4\% versus 93.5\% for Seed-2.0-Lite and 69.0\% versus 86.9\% for
GPT-5.5 when the higher point is nearer versus farther. This separate analysis
also reveals sensitivity to vertical image-plane cues.

\subsubsection{Difficulty analysis}
\label{sec:d3-difficulty}
Tables~\ref{tab:main-difficulty} and~\ref{tab:adaptive-difficulty} separate the 3DCueBench circular results of Tables~\ref{tab:reasoning-main} and~\ref{tab:adaptive-circular}, respectively, by the benchmark's ground-truth relative-gap intervals; smaller gaps indicate more difficult comparisons.

\paragraph{Comparison across models.} In Table~\ref{tab:main-difficulty}, every open-source VLM loses accuracy as the gap shrinks. The strongest, VST-7B-SFT, drops from 88.0\% in the easiest interval to 61.0\% in the hardest, whereas GeoUG drops from 97.0\% to 84.0\%. GeoUG's margin over VST-7B-SFT therefore grows from 9.0 points in the easiest interval to 23.0 points in the hardest. DA3 answering from its own depth follows a similar trend to GeoUG: it is 1.5 and 1.0 points higher in the two hardest intervals, 2.5 points lower in the third, and 0.5 points higher in the easiest.

\paragraph{Draft-conditioned policy.} In Table~\ref{tab:adaptive-difficulty}, the two option-order variants of a question share one generated depth map. Always generating improves over direct answering by 23.0 points in the two hardest intervals and by 12.5 and 5.0 points in the easier ones. The initial policy falls furthest behind always generating in the two hardest intervals (9.5 and 7.0 points). The $\lambda=0.1$ policy stays within 1.0 point of always generating in every interval, and the random gate at the same per-interval rate is 2.0--4.6 points below the policy.

\begin{table}[htbp]
  \centering
  \caption{\textbf{3DCueBench by difficulty: models in Table~\ref{tab:reasoning-main}.} Circular accuracy (\%) in each relative-gap interval (smaller is harder), pooled over Depth and Distance (200 questions per interval).}
  \label{tab:main-difficulty}
  \begingroup
  \setlength{\tabcolsep}{6.8pt}
  \renewcommand{\arraystretch}{1.1}
  \small
  \begin{tabular}{@{}lccccc@{}}
    \toprule
    \multirow{2}{*}{Model / configuration} & \multicolumn{4}{c}{Relative gap $g$} & \multirow{2}{*}{Overall} \\
    \cmidrule(lr){2-5}
    & $[0.10,0.16)$ & $[0.16,0.25)$ & $[0.25,0.40)$ & $[0.40,1]$ & \\
    \midrule
    BAGEL-7B-MoT~\citep{bagel} & 52.0 & 57.0 & 61.5 & 76.0 & 61.6 \\
    Qwen2.5-VL-7B~\citep{qwen25vl} & 42.0 & 45.0 & 51.0 & 56.5 & 48.6 \\
    Qwen3-VL-8B~\citep{qwen3vl} & 54.0 & 61.5 & 71.5 & 80.0 & 66.8 \\
    InternVL3.5-8B~\citep{internvl35} & 48.0 & 52.5 & 60.0 & 70.0 & 57.6 \\
    SenseNova-SI-1.4-8B~\citep{sensenova_si} & 60.5 & 67.0 & 79.0 & 82.0 & 72.1 \\
    SenseNova-SI-1.5-8B~\citep{sensenova_si} & 57.5 & 67.0 & 78.0 & 83.5 & 71.5 \\
    VST-7B-SFT~\citep{vst} & 61.0 & 77.0 & 85.5 & 88.0 & 77.9 \\
    \midrule
    DA3~\citep{da3} (depth readout) & 85.5 & 89.0 & 91.5 & 97.5 & 90.9 \\
    \midrule
    GeoUG (Ours) & 84.0 & 88.0 & 94.0 & 97.0 & 90.8 \\
    \bottomrule
  \end{tabular}
  \endgroup
\end{table}

\begin{table}[htbp]
  \centering
  \caption{\textbf{3DCueBench by difficulty: configurations in Table~\ref{tab:adaptive-circular}.} Circular accuracy (\%), pooled as in Table~\ref{tab:main-difficulty}. The random gate uses the $\lambda=0.1$ policy's rate within each interval.}
  \label{tab:adaptive-difficulty}
  \begingroup
  \setlength{\tabcolsep}{7pt}
  \renewcommand{\arraystretch}{1.1}
  \small
  \begin{tabular}{@{}lccccc@{}}
    \toprule
    \multirow{2}{*}{Model / configuration} & \multicolumn{4}{c}{Relative gap $g$} & \multirow{2}{*}{Overall} \\
    \cmidrule(lr){2-5}
    & $[0.10,0.16)$ & $[0.16,0.25)$ & $[0.25,0.40)$ & $[0.40,1]$ & \\
    \midrule
    Direct & 61.5 & 66.0 & 81.0 & 93.0 & 75.4 \\
    Always-generate & 84.5 & 89.0 & 93.5 & 98.0 & 91.3 \\
    Initial policy (SFT) & 75.0 & 82.0 & 90.0 & 94.5 & 85.4 \\
    \midrule
    Random gate & 80.8 & 84.2 & 89.4 & 95.0 & 86.1 \\
    RL, $\lambda=0.05$ & 83.5 & 87.0 & 94.5 & 97.5 & 90.6 \\
    Ours ($\lambda=0.1$) & 84.0 & 88.0 & 94.0 & 97.0 & 90.8 \\
    \bottomrule
  \end{tabular}
  \endgroup
\end{table}

\FloatBarrier

\section{Additional Experimental Results}
\label{app:additional-results}

\subsection{Generation Pretraining Followed by Understanding SFT}
\label{app:pretraining-sft}

\noindent\textbf{Generation pretraining alone does not reliably provide a better initialization for downstream understanding SFT.} Together with the task-dependent gains from co-training, these results suggest that jointly optimizing generation and understanding may be important for transferring geometric knowledge.

We compare three initializations: the original BAGEL~\citep{bagel} model, a model pretrained for 6,000 steps on depth generation, and a model pretrained for 6,000 steps on multi-view generation. All three then undergo the same understanding-only SFT, using the same data and hyperparameters, with no generation objective during this stage. At steps 1,000--6,000, cumulative SFT CE tokens are 6.29M, 12.48M, 18.77M, 25.00M, 31.10M, and 37.34M for each arm. The preceding generation stage is excluded from the plotted token budget; total training compute is therefore not matched.

\begin{figure}[htbp]
  \centering
  \includegraphics[width=\textwidth]{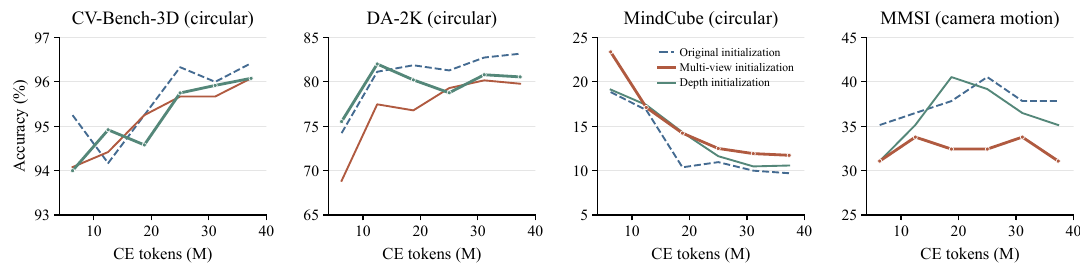}
  \caption{\textbf{Generation pretraining followed by understanding SFT.} Six EMA checkpoints per run, plotted against SFT CE tokens. CV-Bench-3D, DA-2K, and MindCube use circular accuracy; MMSI camera motion uses ordinary accuracy ($n=74$).}
  \label{fig:pretraining-sft}
\end{figure}

At the final SFT checkpoint, both generation-initialized models are 0.34 percentage points below the original initialization on CV-Bench-3D~\citep{rw_cambrian}. Depth and multi-view initialization are lower by 2.61 and 3.38 points on DA-2K~\citep{depthanythingv2}, and by 2.70 and 6.76 points on MMSI~\citep{yang2025mmsibench} camera motion, respectively. MindCube~\citep{wang2025mindcube} shows a different pattern: the generation-initialized models retain small advantages of 0.87 and 2.02 points, but accuracy declines after the first evaluated SFT checkpoint for all three arms. These results concern transfer under the chosen SFT procedure, rather than the quality of the learned representations in isolation. Since the sequential and co-training experiments are not fully matched, they do not establish that co-training is necessary.

\subsection{Visual and Textual Referring Formats}
\label{sec:visual-textual}
We report all seven evaluated open-source models under both visual and textual referring formats. CV-Bench-3D~\citep{rw_cambrian} uses overlaid boxes in its visual version and textual box coordinates in its textual version; both contain the same 1,200 questions (600 Depth and 600 Distance). DA-2K~\citep{depthanythingv2} uses colored point markers in the visual version and textual point coordinates in the textual version, with 2,068 questions. Textual coordinates follow each model's input convention, including conversion to resized-image pixels where required.

\begin{table}[htbp]
\centering\footnotesize
\caption{\textbf{Visual versus textual referring.} Circular accuracy (\%); both variants must be correct. $\Delta$: Textual minus Visual (percentage points). CV-Bench-3D swaps options; DA-2K swaps colors for Visual and options for Textual.}
\label{tab:visual-textual-overall}
\setlength{\tabcolsep}{3pt}
\begin{tabular}{@{}lcccccc@{}}
\toprule
& \multicolumn{3}{c}{CV-Bench-3D} & \multicolumn{3}{c}{DA-2K} \\
\cmidrule(lr){2-4}\cmidrule(l){5-7}
Model & Visual & Textual & $\Delta$ & Visual & Textual & $\Delta$ \\
\midrule
BAGEL-7B-MoT~\citep{bagel} & 81.9 & 70.4 & $-11.5$ & 58.4 & 55.1 & $-3.3$ \\
Qwen2.5-VL-7B~\citep{qwen25vl} & 79.9 & 70.8 & $-9.1$ & 9.8 & 45.3 & $+35.5$ \\
Qwen3-VL-8B~\citep{qwen3vl} & 91.6 & 87.1 & $-4.5$ & 67.8 & 44.6 & $-23.2$ \\
InternVL3.5-8B~\citep{internvl35} & 84.7 & 78.9 & $-5.8$ & 53.7 & 31.6 & $-22.1$ \\
SenseNova-SI-1.4-8B~\citep{sensenova_si} & 87.3 & 89.3 & $+2.0$ & 61.6 & 63.9 & $+2.3$ \\
SenseNova-SI-1.5-8B~\citep{sensenova_si} & 85.8 & 87.4 & $+1.6$ & 65.5 & 61.1 & $-4.4$ \\
VST-7B-SFT~\citep{vst} & 94.9 & 93.2 & $-1.8$ & 75.9 & 63.5 & $-12.4$ \\
\bottomrule
\end{tabular}
\end{table}

\paragraph{Sensitivity varies across models and tasks.}
Textual referring does not uniformly increase or decrease accuracy. On CV-Bench-3D, five models score lower with textual coordinates, while both SenseNova configurations score higher. Referring format therefore affects measured performance and should be held fixed when comparing systems.

\paragraph{DA-2K also probes marker and answer consistency.}
Qwen2.5-VL's 9.8\% visual circular accuracy is accompanied by a strong preference for the green point (93.5\% of responses), while its mean accuracy over visual variants is 53.3\%. Its large Textual-minus-Visual difference therefore cannot be interpreted as an equally large gain in geometric understanding. DA-2K changes both the referring format and the circular transformation; this comparison measures sensitivity to that combined evaluation setup rather than a controlled effect of textual coordinates alone.

\paragraph{Result provenance.}
Textual results use the corrected, model-specific coordinate conventions of the September 2026 baseline evaluation. BAGEL's Visual CV-Bench-3D and DA-2K results are reused from the August 2026 runs. The table reproduces the reported aggregate scores; displayed differences are computed before rounding to one decimal place. Cross-run differences are descriptive comparisons, since equivalence of preprocessing and evaluator versions has not been established.

\subsection{External Depth Tools}
\label{sec:depth-tools}
We also evaluate whether supplying an existing VLM with external depth measurements yields comparable performance. DA3~\citep{da3} (DA3NESTED-GIANT-LARGE-1.1) predicts depth, a shared measurement procedure extracts object depths and pairwise 3D distances, and the measurements are supplied as text to the answering VLM. Table~\ref{tab:depth-tools} reports the resulting system performance alongside the corresponding direct-answer baselines. Qwen2.5-VL uses different input resolutions with and without DA3, and BAGEL prompts also differ; these are reference system comparisons rather than controlled estimates of the tool effect.

\begin{table}[htbp]
  \centering\fontsize{7.5}{9}\selectfont
  \caption{\textbf{External depth tools.} Circular accuracy (\%); both option orders must be correct. All benchmarks use textual referring formats. The last row answers from the DA3 measurements alone, without a VLM, as in Table~\ref{tab:reasoning-main}.}
  \label{tab:depth-tools}
  \setlength{\tabcolsep}{2pt}
  \begin{tabular}{@{}llccc@{}}
    \toprule
    Answering model & Depth tool & CV-Bench-3D & 3DCueBench & DA-2K \\
    \midrule
    BAGEL-7B-MoT~\citep{bagel} & None & 70.4 & 61.6 & 55.1 \\
    BAGEL-7B-MoT~\citep{bagel} & DA3~\citep{da3} & 83.0 & 67.8 & 54.5 \\
    \midrule
    Qwen2.5-VL-7B~\citep{qwen25vl} & None & 70.8 & 48.6 & 45.3 \\
    Qwen2.5-VL-7B~\citep{qwen25vl} & DA3~\citep{da3} & 79.3 & 54.4 & 48.4 \\
    \midrule
    None (depth readout) & DA3~\citep{da3} & 95.3 & 90.9 & 92.1 \\
    \bottomrule
  \end{tabular}
\end{table}

The tool-assisted systems score higher on CV-Bench-3D and 3DCueBench than their corresponding direct-answer baselines, but gains do not hold consistently on DA-2K. Their scores remain below GeoUG and below answering from the DA3 measurements alone (last row) on all three benchmarks. This comparison shows that the evaluated combination of external measurements and an existing VLM is insufficient to match GeoUG; it does not establish that self-generated depth is intrinsically more accurate than DA3 or isolate the effect of tool use from prompting, resolution, and training differences.

\end{document}